\def\isarxiv{1} 

\ifdefined\isarxiv
\documentclass[11pt]{article}
\else
\documentclass{article}
\usepackage{neurips_2022}
\fi

\usepackage{enumitem}
\usepackage{textgreek}
\usepackage{units}
\usepackage{amsmath}
\usepackage{amsthm}
\usepackage{amssymb}
\usepackage{algorithm}
\usepackage{subfig}

\usepackage[english]{babel}
\usepackage{graphicx}
\usepackage{grffile}
\usepackage{wrapfig,epsfig}
\usepackage{epstopdf}
\usepackage{url}
\usepackage{xcolor}
\usepackage{epstopdf}
\usepackage{algpseudocode}
\usepackage[T1]{fontenc}
\usepackage{bbm}
\usepackage{comment}
\usepackage{dsfont}

\usepackage{tikz}
\usepackage{hyperref}  

\ifdefined\isarxiv
\hypersetup{colorlinks=true,citecolor=blue,linkcolor=blue} 
\usetikzlibrary{arrows}
\usepackage[margin=1in]{geometry}
\else 
\definecolor{mydarkblue}{rgb}{0,0.08,0.45}
\hypersetup{colorlinks=true, citecolor=mydarkblue,linkcolor=mydarkblue}
\fi
\graphicspath{{./figs/}}
\usepackage{mathtools}

\newtheorem{theorem}{Theorem}[section]
\newtheorem{lemma}[theorem]{Lemma}
\newtheorem{definition}[theorem]{Definition}

\newtheorem{claim}[theorem]{Claim}

\newcommand{\inner}[2]{\langle #1,#2 \rangle}

\newcommand{\wt}{\widetilde}

\newcommand{\N}{\mathcal{N}}
\newcommand{\R}{\mathbb{R}}

\renewcommand{\varepsilon}{\epsilon}
\renewcommand{\tilde}{\wt}

\DeclareMathOperator*{\argmax}{argmax}
\DeclareMathOperator{\ind}{\mathds{1}}  
\DeclareMathOperator*{\E}{{\mathbb{E}}}

\DeclareMathOperator{\poly}{poly}

\DeclareMathOperator{\tr}{tr}

\DeclareMathOperator{\glo}{global}
\DeclareMathOperator{\loc}{local}

\definecolor{b2}{RGB}{51,153,255}
\definecolor{mygreen}{RGB}{80,180,0}
\definecolor{mycy2}{RGB}{255,51,255}

\makeatletter
\newcommand*{\RN}[1]{\expandafter\@slowromancap\romannumeral #1@}
\makeatother

\begin{document}


\title{Federated Adversarial Learning: A Framework with Convergence Analysis}

\ifdefined\isarxiv

\date{\empty}
\author{
Xiaoxiao Li\thanks{\texttt{xiaoxiao.li@ece.ubc.ca}. University of British Columbia.}
\and
Zhao Song\thanks{\texttt{zsong@adobe.com}. Adobe Research.} 
\and
Jiaming Yang\thanks{\texttt{jiamyang@umich.edu}. University of Michigan, Ann Arbor.}
}

\else
\maketitle

\fi

\ifdefined\isarxiv

\begin{titlepage}
 \maketitle
  \begin{abstract}

Federated learning (FL) is a trending training paradigm to utilize decentralized training data. FL allows clients to update model parameters locally for several epochs, then share them to a global model for aggregation. This training paradigm with multi-local step updating before aggregation exposes unique vulnerabilities to adversarial attacks. Adversarial training is a popular and effective method to improve the robustness of networks against adversaries. In this work, we formulate a \textit{general} form of federated adversarial learning (FAL) that is adapted from adversarial learning in the centralized setting. On the client side of FL training, FAL has an inner loop to generate adversarial samples for adversarial training and an outer loop to update local model parameters. On the server side, FAL aggregates local model updates and broadcast the aggregated model. We design a global robust training loss and formulate FAL training as a min-max optimization problem. Unlike the convergence analysis in classical centralized training that relies on the gradient direction, it is significantly harder to analyze the convergence in FAL for three reasons: 1) the complexity of min-max optimization, 2) model not updating in the gradient direction due to the multi-local updates on the client-side before aggregation and 3) inter-client heterogeneity. We address these challenges by using appropriate gradient approximation and coupling techniques and present the convergence analysis in the over-parameterized regime. Our main result theoretically shows that the minimum loss under our algorithm can converge to $\epsilon$ small with chosen learning rate and communication rounds. It is noteworthy that our analysis is feasible for non-IID clients.

  \end{abstract}
 \thispagestyle{empty}
 \end{titlepage}

\newpage
{
\tableofcontents
}
\newpage

\else
  \begin{abstract}

Federated learning (FL) is a trending training paradigm to utilize decentralized training data. FL allows clients to update model parameters locally for several epochs, then share them to a global model for aggregation. This training paradigm with multi-local step updating before aggregation exposes unique vulnerabilities to adversarial attacks. Adversarial training is a popular and effective method to improve the robustness of networks against adversaries. In this work, we formulate a \textit{general} form of federated adversarial learning (FAL) that is adapted from adversarial learning in the centralized setting. On the client side of FL training, FAL has an inner loop to generate adversarial samples for adversarial training and an outer loop to update local model parameters. On the server side, FAL aggregates local model updates and broadcast the aggregated model. We design a global robust training loss and formulate FAL training as a min-max optimization problem. Unlike the convergence analysis in classical centralized training that relies on the gradient direction, it is significantly harder to analyze the convergence in FAL for three reasons: 1) the complexity of min-max optimization, 2) model not updating in the gradient direction due to the multi-local updates on the client-side before aggregation and 3) inter-client heterogeneity. We address these challenges by using appropriate gradient approximation and coupling techniques and present the convergence analysis in the over-parameterized regime. Our main result theoretically shows that the minimum loss under our algorithm can converge to $\epsilon$ small with chosen learning rate and communication rounds. It is noteworthy that our analysis is feasible for non-IID clients.

  \end{abstract}

\fi

\section{Introduction}
\label{sec:intro}

Federated learning (FL) is playing an important role nowadays, as it allows different clients to train models collaboratively without sharing private information. One popular FL paradigm called FedAvg~\cite{mmr+17} introduces an easy-to-implement distributed learning method without data sharing. Specifically, it requires a central server to aggregate model updates computed by the local clients (also known as nodes or participants) using local imparticipable private data. Then with these updates aggregated, the central server use them to train a global model.

 Nowadays deep learning model are exposed to severe threats of adversarial samples. Namely, small adversarial perturbations on the inputs will dramatically change the outputs or output wrong answers~\cite{szsbegf13}. In this regard, much effort has been made to improve neural networks' resistance to such perturbations using adversarial learning~\cite{tkp+17,skc18,mmstv18}. Among these studies, the adversarial training scheme in \cite{mmstv18} has achieved the good robustness in practice. \cite{mmstv18} proposes an adversarial training scheme that uses projected gradient descent (PGD) to generate alternative adversarial samples as the augmented training set. Generating adversarial examples during neural network training is considered as one of the most effective approaches for adversarial training up to now according to the literature~\cite{cw17c,acw18,ch20}. 

Although adversarial learning has attracted much attention in the centralized domain, its practice in FL is under-explored~\cite{zrsb20}. Like training classical deep neural networks that use gradient-based methods,  FL paradigms are vulnerable to adversarial samples. Adversarial learning in FL brings multiple open challenges due to FL properties on low convergence rate, application in non-IID environments, and secure aggregation solutions. Hence applying adversarial learning in an FL paradigm may lead to unstable training loss and a lack of robustness. However, a recent practical work~\cite{zrsb20} observed that although there exist difficulties of convergence, the federation of adversarial training with suitable hyperparameter settings can achieve adversarial robustness and acceptable performance. Motivated by the empirical results, we want to address the provable property of combining adversarial learning into FL from the theoretical perspective. 

This work aims to theoretically study the unexplored convergence challenges that lie in the interaction between adversarial training and FL. To achieve a general understanding, we consider a general form of \textbf{\textit{federated adversarial learning (FAL)}}, which deploys adversarial training scheme on local clients in the most common FL paradigm, FedAvg~\cite{mmr+17} system. Specifically, FAL has an inner loop of local updating that generates adversarial samples (i.e., using ~\cite{mmstv18}) for adversarial training and an outer loop to update local model weights on the client side. Then global model is aggregated using FedAvg~\cite{mmr+17}. Our algorithm is detailed in  Algorithm~\ref{alg:fl_adv_train_main:intro}.

We are interested in theoretically understanding the proposed FAL scheme from the aspects of model robustness and convergence:
\begin{center}
    \emph{Can federated adversarial training fit training data robustly and converge with an over-parameterized neural network?}
\end{center}

The theoretical convergence analysis of adversarial training itself is challenging in the centralized training setting. \cite{tzt18} recently proposed a general theoretical method to analyze the risk bound with adversaries but did not address the convergence problem. The investigation of convergence on over-parameterized neural network has achieved tremendous progress~\cite{dllwz19,als19a,als19b,dzps19,adhlw19}. The basic statement is that training can converge to sufficiently small training loss in polynomial iterations using gradient descent or stochastic gradient descent when the width of the network is polynomial in the number of training examples when initialized randomly.  Recent theoretical analysis~\cite{gcl+19,zpdlsa20} extends these standard training convergence results to adversarial training settings. To answer the above interesting but challenging question, we formulate FAL as an min-max optimization problem. We extend the convergence analysis on the general formulation of over-parameterized neural networks in the FL setting that allows each client to perform min-max training and generate adversarial examples (see Algorithm~\ref{alg:fl_adv_train_main:intro}). Involved challenges are arising in FL convergence analysis due to its unique optimization method: 1) unlike classical centralized setting, the global model of FL does not update in the gradient direction; 2) inter-client heterogeneity issue needs to be considered.

Despite the challenges, we give an affirmative answer to the above question. To the best of our knowledge, this work is the first theoretical study that studies those unexplored problems about the convergence of adversarial training with FL. The contributions of this paper are:

\begin{itemize}
    \item We propose a framework to analyze a general form of FAL in over-parameterized neural networks. 
    We follow a natural and valid assumption of data separability that the training dataset are well separated apropos of the adversarial perturbations' magnitude. After sufficient rounds of global communication and certain steps of local gradient descent for each $t$, we obtain the minimal loss close to zero. Notably, our assumptions do not rely on data distribution. Thus the proposed analysis framework is feasible for non-IID clients.
    \item We are the first to theoretically formulate the convergence of the FAL problem into a min-max optimization framework with the proposed loss function. In FL, the update in the global model is no longer directly determined by the gradient directions due to multiple local steps. To tackle the challenges, we define a new `gradient', \textit{FL gradient}. With valid ReLU Lipschitz and over-parameterized assumptions, we use gradient coupling for gradient updates in FL to show the model updates of each global updating is bounded in federated adversarial learning.
\end{itemize}

\section{Related Work}
\label{sec:related}

\paragraph{Federated Learning}
 
 A efficient and privacy-preserving way to learn from the distributed data collected on the edge devices (a.k.a clients) would be FL.  
 FedAvg is a easy-to-implement distributed learning strategy by aggregating local model updates of the server's side, and then transmitting the averaged model back to the local clients. Later, many FL methods are developed baed on FedAvg. Theses FL schemes can be divided into aggregation schemes~\cite{mmr+17,wll+20,ljz+21} and optimization schemes~\cite{rcz+20,zhd+20}.  Nearly all the them have the common characteristics that client model are updating using gradient descent-based methods, which is venerable to adversarial attacks. In addition, data heterogeneity brings in huge challeng in FL. For IID data, FL has been proven effective. However, in practice, data mostly distribute as non-IID. Non-IID data could substantially degrade the performance of FL models~\cite{zls+18,lhy+19,ljz+21,lsz+20}. Despite the potential risk in security and unstable performance in non-IID setting, as FL mitigates the concern of data sharing, it is still a popular and practical solution for distributed data learning in many real applications, such as  healthcare~\cite{lgd+20,rhl+20}, autonomous driving~\cite{llc+19}, IoTs~\cite{wts+19,llh+20}.

\paragraph{Learning with Adversaries}
Ever since adversarial examples are discovered~\cite{szsbegf13}, to make neural networks robust to perturbations, efforts have been made to propose more effective defense methods. As adversarial examples are an issue of robustness, the popular scheme is to include learning with adversarial examples, which can be traced back to \cite{gss14}. It produces adversarial examples and injecting them into training data. Later, Madry
et al.~\cite{mmstv18} proposed training on multi-step PGD adversaries and empirically observed that adversarial training consistently achieves small and robust training loss in wide neural networks.

\paragraph{Federated Adversarial Learning} 
Adversarial examples, which may not be visually distinguishable from benign samples, are often classified. This poses potential security threats for practical machine learning applications. Adversarial training~\cite{gss14,kgb16b} is a popular protocol to train more adversarial robust models by inserting adversarial examples in training. The use of adversarial training in FL presents a number of open challenges, including poor convergence due to multiple local update steps, instability and heterogeneity of clients, cost and security request of communication, and so on. To defend the adversarial attacks in federated learning, limited recent studies have proposed to include adversarial training on clients in the local training steps ~\cite{bcmc19,zrsb20}. These two works empirically showed the performance of adversarial training, while the theoretical analysis of convergence is under explored. 
\cite{dkm20} focused the problem of distributionally robust FL with an emphasis on reducing the communication rounds, they traded $O(T^{1/8})$ convergence rate for $O(T^{1/4})$ communication rounds.
In addition, different from our focus on a generic theoretical analysis framework, \cite{zll+21} is a methodology paper that proposed an adversarial training strategy in classical distributed setting, with focus on specific training strategy (PGD, FGSM), which could be generalized to a method in FAL.

\paragraph{Convergence via Over-parameterization} Convergence analysis on over-parameterized neural networks falls in two lines. In the first line of work \cite{ll18,als19a,als19b,all19} data separability plays a crucial role, and is widely used in theoretically showing the convergence result in the over-parameterized neural network setting. To be specific, data separability theory shows that to guarantee convergence, the width ($m$) of a neural network shall be at least polynomial factor of all parameters (i.e. $m \geq \poly(n,d,1/\delta)$), where $\delta$ is the minimum distance between all pairs of data points, $n$ is the number of data points and $d$ is the data dimension. 
Another line of work \cite{dzps19,adhlsw19,adhlw19,sy19,lsswy20,bpsw21,syz21,szz21,hlsy21,z22,mosw22} builds on neural tangent kernel (NTK) \cite{jgh18}. In this line of work, the minimum eigenvalue ($\lambda$) of the NTK is required to be lower bounded to guarantee convergence. Our analysis focuses on the former approach based on data separability.

\paragraph{Robustness of Federated Learning}
Previously there were several works that theoretically analyzed the robustness of federated learning under noise. \cite{yckb18} developed distributed optimization algorithms that were provably robust against arbitrary and potentially adversarial behavior in distributed computing systems, and mainly focused on achieving optimal statistical performance. \cite{rfpj20} developed a robust federated learning algorithm by considering a structured affine distribution shift in users’ data. Their analysis was built on several assumptions on the loss functions without a direct connection to neural network.

\section{Problem Formulation}\label{sec:fromulation_main}
To explore the properties of FAL in deep learning, we formulate the problem in over-parameterized neural network regime. We start by presenting the notations and setup required for federated adversarial learning, then we will describe the loss function we use and our FAL algorithm.

\subsection{Notations}\label{sec:3.1}
For a vector $x$, we use $\| x\|_p$ to denote its $\ell_p$ norm, in this paper we mainly consider the situation when $p=1,2,$ or $\infty$. For a matrix $U \in \R^{d \times m}$, we use $U^\top$ to denote its transpose and use $\tr[U]$ to denote its trace. We use $\|U\|_1$ to denote its entry-wise $\ell_1$ norm. We use $\|U\|_2$ to denote its spectral norm. We use $\|U\|_F$ to denote its Frobenius norm. For $j \in [m]$, we let $U_j \in \R^d$ be the $j$-th column of $U$. We let $\| U \|_{2,1}$ denotes $\sum_{j=1}^m \|U_j \|_2$. We let $\| U \|_{2,\infty}$ denotes $\max_{j \in [m]} \| U_j \|_2$. 

We denote Gaussian distribution with mean $\mu$ and covariance $\Sigma$ as ${\cal N}(\mu,\Sigma)$. We use $\sigma(\cdot)$ to denote the ReLU function $\sigma(x)=\max\{x,0\}$, and use $\ind\{A\}$ to denote the indicator function of  event $A$.

\subsection{Problem Setup}\label{sec:3.2}
\paragraph{Two-layer ReLU network in FAL}

Following recent theoretical work in understanding neural networks training in deep learning~\cite{dzps19,adhlsw19,adhlw19,sy19,lsswy20,syz21,z22}, in this paper, we focus on a two-layer neural network that has $m$ neurons in the hidden layer, where each neuron is a ReLU activation function.  

We define the global network as  
\begin{align}\label{eq:def_f_U:intro}
f_{U}(x) := \sum_{r=1}^m a_r  \cdot \sigma ( \langle U_{r},x \rangle+b_r )
\end{align}
and for $c\in [N]$, we define the local network of client $c$ as
\begin{align}\label{eq:def_f_W:intro}
    f_{W_c}(x) := \sum_{r=1}^m a_r  \cdot \sigma ( \langle W_{c,r},x \rangle+b_r ).
\end{align}
Here $U = (U_{1}, U_2, \dots, U_{m})\in \mathbb{R}^{d\times m}$ is the global hidden weight matrix, $W_c = (W_{c,1},\dots, W_{c,m})\in \mathbb{R}^{d\times m}$ is the local hidden weight matrix of client $c$, $a=(a_1,a_2,\dots,a_m)\in \mathbb{R}^m$ denotes the output weight, $b=(b_1,b_2,\dots,b_m)\in \mathbb{R}^m$ denotes
the bias.

During the process of federated adversarial learning, we only update the value of $U$ and $W$, while keeping $a$ and $b$ equal to their initialization, so we can write the global network as $f_U(x)$ and the local network as $f_{W_c}(x)$. For the situation we don't care about the weight matrix, we write $f(x)$ or $f_c(x)$ for short.

Next, we make some standard assumptions regarding our training set.

\begin{definition}[Dataset]
There are $N$ clients and $n=N J$ data in total.\footnote{For simplicity, we assume that all clients have same number of training data. Our result can be generalized to the setting where each client has a different number of data as the future work.} Let $\mathcal{S} = \cup_{c \in [N] } \mathcal{S}_c$ where  $\mathcal{S}_c= \{(x_{c,1},y_{c,1}),...,(x_{c,J},y_{c,J}) \} \subseteq \mathbb{R}^d\times \mathbb{R}$ denotes the $J$ training data of client $c$. Without loss of generality, we assume $\|x_{c,j}\|_2=1$ holds for all $c\in[N], j\in[J]$, and the last coordinate of each point equals to $1/2$
, so we consider ${\cal X} : = \{ x \in \R^d : \| x \|_2 =1 ,\ x_{d} = 1/2 \}$. For simplicity, we assume that $|y_{c,j}|\leq 1$ holds for all $c\in[N]$ and $j\in[J]$.\footnote{Our assumptions on data points are reasonable since we can do scale-up. In addition, $l_2$ norm normalization is a typical technique in experiments. Same assumptions also appears in many previous theoretical works like \cite{adhlw19,all19,als19a}.}
\end{definition}

We now define the initialization for the neural networks. 
\begin{definition}[Initialization]\label{def:initialize}
The initialization of $a \in \R^m, U \in \R^{d \times m}, b \in \R^m$ is $a(0) \in \R^m, U(0) \in \R^{d \times m}, b(0) \in \R^m$. The initialization of client c's local weight matrix $W_c$ is $W_c(0,0)=U(0)$. Here the second term in $W_c$ denotes iteration of local steps.
\begin{itemize}
    \item For each $r\in[m]$, $a_r(0)$ are i.i.d. sampled from $ [- 1/ m^{1/3}, +1/m^{1/3}]$ uniformly.
    \item For each $i\in[d], r\in[m]$, $U_{i,r}(0)$ and $b_r(0)$ are i.i.d. random Gaussians sampled from $\mathcal{N}(0, 1/m )$. Here $U_{i,r}$ means the $(i,r)$-entry of $U$.
\end{itemize}
For each global iteration $t\in[T]$,
\begin{itemize}
    \item For each $c\in[N]$, the initial value of client c's local weight matrix $W_c$ is $W_c(t,0)=U(t)$.
\end{itemize}
\end{definition}

Next we formulate the adversary model that will be used.
\begin{definition}[$\rho$-Bounded adversary]  
Let $\mathcal{F}$ denote the function class. An adversary is a mapping $\mathcal{A}: \mathcal{F} \times \mathcal{X} \times \R \rightarrow \mathcal{X}$ which denotes the adversarial perturbation. For $\rho>0$, we define the $\ell_2$ ball ${\cal B}_2(x,\rho) := \{ \wt{ x}\in \R^d : \| \wt{x} - x\|_2 \leq \rho \}\cap \cal{X}$, we say an adversary $\mathcal{A}$ is \textbf{$\rho$-bounded} if it satisfies $\mathcal{A}(f, x, y) \in \mathcal{B}_2(x,\rho)$. Furthermore, given $\rho >0$, we denote the \textbf{worst-case} adversary as $\mathcal{A}^{*} := \argmax_{\tilde{x}\in\mathcal{B}_2(x,\rho)} \ell(f(\tilde{x}), y)$, where $\ell$ is defined in Definition~\ref{def:loss_l}.
 
\end{definition}

\paragraph{Well-separated training set}
In the over-parameterized regime, it is a standard assumption that the training set is well-separated. Since we deal with adversarial perturbations, we require the following $\gamma$-separability, which is a bit stronger.

\begin{definition}[$\gamma$-separability] \label{def:sep}
Let $\gamma \in (0,1/2), \delta \in (0,1/2), \rho \in (0,1/2)$ denote three parameters such that $\gamma \leq \delta \cdot ( \delta - 2 \rho )$.
We say our training set $\mathcal{S} = \cup_{c \in [N] } \mathcal{S}_c = \cup_{c \in [N], j\in[J] }\{ (x_{c,j}, y_{c,j})\} \subset \R^d \times \R$ is \textbf{globally $\gamma$-separable} w.r.t a $\rho$-bounded adversary, if $\| x_{c_1,j_1} - x_{c_2,j_2} \|_2 \geq \delta$ holds for any $c_1\neq c_2$ and $j_1\neq j_2$.

\end{definition}
Note that in the above definition, the introducing of $\gamma$ is for expression simplicity of Theorem~\ref{thm:main}, and the assumption $\gamma \leq \delta \cdot (\delta - 2 \rho)$ is reasonable and easy to achieve in adversarial training.
It is also noteworthy that, our problem setup does not need the assumption on independent and identically distribution (IID) on data, thus such a formation can be applied to unique challenge of the non-IID setting in FL.

\subsection{Federated Adversarial Learning}\label{sec:3.3}

\paragraph{Adversary and robust loss}

We set the following loss for the sake of technical presentation simplicity, as is customary in prior studies~\cite{gcl+19, all19}:

\begin{definition}[Lipschitz convex loss]\label{def:loss_l}
A loss function $\ell: \R\times \R \rightarrow \R$ is said to be a \textbf{Lipschitz convex loss}, if it satisfies the following properties: (i) convex w.r.t. the first input of $\ell$; (ii) $1-$Lipshcitz, which means $| \ell(x_1, y_1) -\ell(x_2,y_2)| \leq \|(x_1,y_1)-(x_2,y_2)\|_2;$ and (iii) $\ell(y, y) = 0$ for all $ y\in\R$.
\end{definition}
 
In this paper we assume $\ell$ is a Lipschitz convex loss. Next we define our robust loss function of a network, which is based on the adversarial samples generated by a $\rho$-bounded adversary $\mathcal{A}$. 

\begin{definition}[Training loss]\label{def:loss}
Given a client's training set $\mathcal{S}_c = \{ (x_{c,j}, y_{c,j})\}_{j=1}^J \subset \R^d \times \R$ of $J$ samples. Let $f_c : \R^d \rightarrow \R$ be a net. The classical training loss of $f_c$ is 
$
\mathcal{L}(f_c, S_c) := \frac{1}{J} \sum_{j=1}^J \ell\left(f_c(x_{c,j}), y_{c,j}\right)
$. 
Given $\mathcal{S} = \cup_{c \in [N] } \mathcal{S}_c$, we define the global loss as 
\begin{align*} 
\mathcal{L}(f_U,S) := \frac{1}{NJ}\sum_{c=1}^N\sum_{j=1}^J \ell(f_U(x_{c,j}),y_{c,j}).
\end{align*}
 
Given an adversary $\mathcal{A}$ that is $\rho$-bounded, we define
 
the global loss with respect to $\mathcal{A}$ as
\begin{align*}
    {\cal L}_\mathcal{A}(f_{U}) := \frac{1}{NJ}\sum_{c=1}^N\sum_{j=1}^J \ell (f_U(\mathcal{A}(f_c, x_{c,j},y_{c,j})), y_{c,j}) = \frac{1}{NJ}\sum_{c=1}^N\sum_{j=1}^J \ell (f_U(\tilde{x}_{c,j}), y_{c,j})
\end{align*}
 
and also define
the global robust loss (in terms of \textbf{worst-case}) as
\begin{align*}
    {\cal L}_{\mathcal{A}^*}(f_{U}) := \frac{1}{NJ}\sum_{c=1}^N\sum_{j=1}^J \ell (f_U(\mathcal{A}^*(f_c, x_{c,j},y_{c,j})), y_{c,j}) = \frac{1}{NJ}\sum_{c=1}^N\sum_{j=1}^J \max_{{x}_{c,j}^*\in\mathcal{B}_2(x_{c,j},\rho)}\ell\left(f_U({x}_{c,j}^*), y_{c,j}\right).
\end{align*}
 
Moreover, since we deal with pseudo-net (Definition~\ref{def:pseudo:intro}), we also define the loss of a pseudo-net as
\begin{align*}
    \mathcal{L}(g_c, \mathcal{S}_c) :=\frac{1}{J} \sum_{j=1}^J \ell\left(g_c(x_{c,j}), y_{c,j}\right)
\end{align*}
and
\begin{align*}
    \mathcal{L}(g_{U},\mathcal{S}) :=\frac{1}{NJ}\sum_{c=1}^N\sum_{j=1}^J \ell(g_U(x_{c,j}),y_{c,j}).
\end{align*}
\end{definition}

\paragraph{Algorithm}
We focus on a general FAL framework that is adapted from the most common adversarial training in the classical setting on the client. Specifically, we describe the adversarial learning of a local neural network $f_{W_c}$ against an adversary $\mathcal{A}$ that generate adversarial examples during training as shown in Algorithm~\ref{alg:fl_adv_train_main:intro}. As for the analysis of a general theoretical analysis framework, we do not specify the explicit format of $\mathcal{A}$. 

The FAL algorithm contains two procedures: one is \textsc{ClientUpdate} running on client side and the other is \textsc{ServerExecution} running on server side. These two procedures are iteratively processed through communication iterations. Adversarial training is addressed in procedure \textsc{ClientUpdate}. Hence, there are two loops in \textsc{ClientUpdate} procedure: the outer loop is iteration for local model updating; and the inner loop is iteratively generating adversarial samples by the adversary $\mathcal{A}$. In the outer loop in \textsc{ServerExecution} procedure, the neural network's parameters are updated 
to reduce its prediction loss on the new adversarial samples.

\begin{algorithm*}[!ht]
    \caption{Federated Adversarial Learning (FAL)}
    \label{alg:fl_adv_train_main:intro}
    \textbf{Notations: }{Training sets of clients with each client is indexed by $c$, $\mathcal{S}_c=\{(x_{c,j},y_{c,j})\}_{j=1}^J$; adversary $\mathcal{A}$; local learning rate $\eta_{\loc}$; global learning rate $\eta_{\mathrm{global}}$; local updating iterations $K$; global communication round $T$.}
	\begin{algorithmic}[1]
	\State{Initialization $a(0) \in \R^m, U(0) \in \R^{d \times m}, b(0) \in \R^m$}
	\State{For $t = 0 \to T$, we iteratively run \textbf{Procedure A} then \textbf{Procedure B}}
	\Procedure{\textbf{A}. ClientUpdate}{$t,c$} 
	            \State $\mathcal{S}_c(t) \leftarrow \emptyset$
	            \State $W_{c}(t, 0)\leftarrow U(t)$ \Comment{Receive global model weights update.}
	            \For{$k=0 \to K-1$ }
	                \For{$j = 1 \to J$}
	                    \State $\tilde{x}_{c,j}^{(t)} \leftarrow \mathcal{A}(f_{W_{c}(t,k)}, x_{c,j}, y_{c,j} )$ \Comment{Adversarial samples. $f_{W_c}$ is defined as \eqref{eq:def_f_W:intro}.}  
	                    \State $\mathcal{S}_c(t) \leftarrow \mathcal{S}_c(t) \cup (\tilde{x}_{c,j}^{(t)}, y_{c,j})$

	                \EndFor
	                \State $W_{c}{(t,k+1)} \leftarrow W_{c}{(t,k)} - \eta_{\loc} \cdot~\nabla_{W_c} \mathcal{L}(f_{W_{c}(t,k)}, \mathcal{S}_c(t) )$
	                \EndFor
	            \State $\Delta U_c(t) \leftarrow W_{c}(t,K) - U(t)$
	    \State Send $\Delta U_c(t)$ to \textsc{ServerExecution}
	\EndProcedure
	
    \Procedure{\textbf{B}. ServerExecution} {$t$}:
                \For{each client $c$ \textbf{in parallel do}}
                \State{$\Delta U_c(t) \gets$ \textsc{ClientUpdate}$(c,t)$}
                \Comment{Receive local model weights update.}
  	        \State $\Delta U(t) \leftarrow \frac{1}{N} \sum_{c \in [N]} \Delta U_c(t)$  
	        \State $U(t+1) \leftarrow U(t) + \eta_{\mathrm{global}} \cdot \Delta U(t)$
                \Comment{Aggregation on the server side.}
                \State{Send $U(t+1)$ to client $c$ for \textsc{ClientUpdate}$(c,t)$}
        \EndFor
        \EndProcedure

	\end{algorithmic}
\end{algorithm*}

\section{Our Result}\label{sec:results}

The main result of this work is showing the convergence of FAL algorithm (Algorithm~\ref{alg:fl_adv_train_main:intro}) in overparameterized neural networks. Specifically, our defined global training loss (Definition~\ref{def:loss})  converges to a small $\epsilon$ with the chosen communication round $T$, local and global learning rate $\eta_{\loc}$, $\eta_{\glo}$. 
 
We now formally present our main result.
\begin{theorem}[Federated Adversarial Learning]\label{thm:main}
Let $c_0 \in (0,1)$ be a fixed constant. 
Let $N$ denotes the total number of clients and $J$ denotes the number of data points per client. Suppose that our training set $\mathcal{S} = \cup_{c \in [N] } \mathcal{S}_c$ 
is globally $\gamma$-separable for some $\gamma>0$. Then, for all $\epsilon\in(0,1)$, there exists $R=\poly(( NJ/\epsilon )^{1/\gamma})
$ that satisfies: for every $K\geq1$ and $T\geq\poly(R/\epsilon)$, for all $m\geq \poly (d, ( NJ/\epsilon )^{1/\gamma} )$, with probability $ \geq 1- \exp( -\Omega( m^{1/3} ) )$ , running \textbf{federated adversarial learning} (Algorithm~\ref{alg:fl_adv_train_main:intro}) with step size choices
 
\begin{align*}
    \eta_{\glo}= 1/ \poly(NJ,R,1/\epsilon)~~~ and~~~\eta_{\loc} = 1/K
\end{align*}
will output a list of weights $\{ U(1), U(2) , \cdots, U(T) \} \in \R^{d \times m}$ that satisfy:

\begin{align*}
    \min_{t \in [T]} \mathcal{L}_{\mathcal{A}}(f_{ U(t) }) \leq \epsilon.
\end{align*}
The randomness comes from $a(\tau) \in \R^m$, $U(\tau) \in \R^{d \times m}$, $b(\tau) \in \R^m $ for $\tau = 0$.
\end{theorem}

\paragraph{Discussion}
As we can see in Theorem~\ref{thm:main}, one key element that affects parameters $m, R, T$ is the data separability $\gamma$. As the data separability bound becomes larger, the parameter $R$ becomes smaller, resulting in the need of a larger global learning rate $\eta_{\glo}$ to achieve convergence. We also conduct numerical experiments in Appendix~\ref{sec:numerical_result} to verify Theorem~\ref{thm:main} empirically.
\section{Proof Sketch}
\label{sec:sketch_main}

To handle the min-max objective in FAL, we formulate the optimization of FAL in the framework of \emph{online gradient descent}\footnote{We refer our readers to \cite{h16} for more details regarding online gradient descent.} : at each local step $k$ on the client side, firstly the adversary generates adversarial samples and computes the loss function $\mathcal{L}\left(f_{W_c(t,k)}, \mathcal{S}_c(t)\right)$, then the local client learner takes  the fresh loss function and update $W_c(t,k+1) = W_c(t,k) - \eta_{\loc} \cdot \nabla_{W_c} \mathcal{L}\left(f_{W_c(t,k)}, \mathcal{S}_c(t)\right)$.

Compared with the centralized setting, the key difficulties in the convergence analysis of FL are induced by multiple local step updates of the client side and the step updates on both local and global sides. Specifically, local updates are not the standard gradient as the centralized adversarial training when $K \geq 2$. We used $-\Delta U(t)$ in substitution of the real gradient of $U$ to update the value of $U(t)$. This brings in challenges to bound the gradient of the neural networks. Nevertheless, gradient bounding is challenging in adversarial training solely.

To this end, we use gradient coupling method twice to solve this core problem: firstly we bound the difference between real gradient and FL gradient (defined below), then we bound the difference between pseudo gradient and real gradient.

\subsection{Existence of small robust loss}\label{sec:E_def_main}

In this section, we denote $\tilde{U}=U(0)$ as the initialization of global weights $U$ and denote $U(t)$ as the global weights of communication round $t$. $U^*$ is the value of $U$ after small perturbations from $\tilde{U}$ which satisfies $\|U^*-\tilde{U}\|_{2,\infty}\leq R/m^{c_1}$, here $c_1\in(0,1)$ is a constant (e.g. $c_1 = 2/3$), $m$ is the width of the neural network and $R$ is a parameter. We will specify the concrete value of these parameters later in appendix.

We study the over-parameterized neural nets' well-approximated pseudo-network to learn gradient descent for over-parameterized neural nets whose weights are close to initialization. Pseudo-network can be seen as a linear approximation of our two layer ReLU neural network near initialization, and the introducing of pseudo-network makes the proof more intuitive.

\begin{definition}[Pseudo-network]\label{def:pseudo:intro}
Given weights $U \in \R^{d \times m}$, $a \in \R^m$ and $b \in \R^{m}$, for a neural network $f_U(x) = \sum_{r=1}^{m}a_r\cdot \sigma(\langle U_r, x \rangle+b_r)$, we define the corresponding \textbf{pseudo-network} $g_U : \R^d \rightarrow \R$ as
$
g_{U}(x) := \sum_{r=1}^m a_r \cdot \langle U_r(t) - U_r(0) , x \rangle \cdot \ind \{\langle U_r(0), x \rangle + b_r \geq 0 \}.
$

\end{definition}

\paragraph{Existence of small robust loss}
To obtain our main theorem, first we show that we can find a $U^*$ which is close to $U(0)$ and also makes $\mathcal{L}_{\mathcal{A}^*}(f_{U^*})$ sufficiently small. Later in Theorem~\ref{thm:convergence:intro} we show that the average of $\mathcal{L}_{\mathcal{A}}(f_{U(t)})$ is dominated by $\mathcal{L}_{\mathcal{A}^*}(f_{U^*})$, thus we can prove the minimum of $\mathcal{L}_{\mathcal{A}}(f_{U(t)})$ is $\epsilon$ small.

\begin{theorem}[Existence, informal version of Theorem~\ref{thm:existence}]\label{thm:existence:intro} 
For all $\epsilon\in(0,1)$, there are $M_0 = \poly (d, ( NJ / \epsilon )^{1/\gamma} )$ and $R = \poly( ( NJ / \epsilon )^{1/\gamma} )$ satisfying: for all $m \geq M_0$, with high probability 
 there exists $U^*\in\mathbb{R}^{d\times m}$ that satisfies $\|U^*-U(0)\|_{2,\infty} \leq R / m^{c_1}$ and $\mathcal{L}_{\mathcal{A}^*} ( f_{U^*} ) \leq \epsilon$.
\end{theorem}

\subsection{Convergence result for federated learning}
 
\begin{definition}[Gradient]\label{def:gradient:intro}
For a local real network $f_{W_c(t,k)}$, we denote its gradient by 
\begin{align*}
    \nabla (f_c,t,k) :=  \nabla_{W_c}\mathcal{L}(f_{W_c(t,k)}, \mathcal{S}_c(t)).
\end{align*}
If the corresponding pseudo-network is $g_{W_c(t,k)}$, then denote the pseudo-network gradient by  
\begin{align*}
    \nabla (g_c,t,k) :=   \nabla_{W_c}\mathcal{L}(g_{W_c(t,k)}, \mathcal{S}_c(t)).
\end{align*}
\end{definition}

Now we consider the global network. We define pseudo gradient as $\nabla(g,t) := \nabla_U\mathcal{L}(g_{U(t)}, \mathcal{S}(t))$ and define \textbf{FL gradient} as $\wt{\nabla}(f,t) := -\frac{1}{N} \Delta U(t)$, which is used in the proof of Theorem~\ref{thm:convergence:intro}. We present our gradient coupling methods in the following two lemmas.
 
\begin{lemma}[Bound the difference between real gradient and FL gradient, informal version of Lemma~\ref{lemma:delta_gradient}]\label{lemma:delta_gradient:intro}
With probability $\geq 1-\exp(-\Omega(m^{c_0}))$,
for iterations $t$ satisfying $\| U(t) - U(0)\|_{2,\infty} \leq 1/o(m)$, the gradients satisfy
\begin{align*}
    \| \nabla (f, t) - \wt{\nabla}(f,t) \|_{2,1} \leq o(m).
\end{align*}
The randomness is from $a(\tau) \in \R^m$, $U(\tau) \in \R^{d \times m}$, $b(\tau) \in \R^m $ for $\tau = 0$.
\end{lemma}

\begin{lemma}[Bound the difference between pseudo gradient and real gradient, informal version of Lemma~\ref{lemma:coupling}]  
\label{lemma:coupling:intro}
With probability $\geq 1-\exp(-\Omega(m^{c_0}))$, for iterations $t$ satisfying $\| U(t) - U(0)\|_{2,\infty} \leq 1/o(m)$, the gradients satisfy  
\begin{align*}
  \| \nabla (g, t) - \nabla(f,t) \|_{2,1} \lesssim NJ\cdot o(m).
\end{align*}
The randomness is from $a(\tau) \in \R^m$, $U(\tau) \in \R^{d \times m}$, $b(\tau) \in \R^m $ for $\tau = 0$.
\end{lemma}
The above two lemmas are essential in proving Theorem~\ref{thm:convergence:intro}, which is our convergence result.

\begin{theorem}[Convergence result, informal version of Theorem~\ref{thm:convergence}]\label{thm:convergence:intro}
Let $R\geq 1$. Suppose $\epsilon\in(0,1)$. Let $K \geq 1$, let $T\geq \poly(R/\epsilon)$. There is $M = \poly( n, R, 1/\epsilon )$, such that for every $m\geq M$, with probability $\geq 1- \exp ( -\Omega (m^{c_0} ) )$,  
for every $U^*$ satisfying $\| U^* - U(0) \|_{2,\infty} \leq R/m^{c_1}$, running Algorithm~\ref{alg:fl_adv_train_main:intro} with setting
$
\eta_{\glo}= 1/ \poly(NJ,R,1/\epsilon)$  and $\eta_{\loc} = 1/K
$ will output weights $(U(t))_{t=1}^T$ that satisfy 
\begin{align*}
\frac{1}{T}\sum_{t=1}^T \mathcal{L}_{\mathcal{A}}\left(f_{U(t)}\right) \leq\mathcal{L}_{\mathcal{A}^*}\left(f_{U^*}\right) + \epsilon.
\end{align*}
The randomness comes from $a(\tau) \in \R^m$, $U(\tau) \in \R^{d \times m}$, $b(\tau) \in \R^m $ for $\tau = 0$.
\end{theorem}
In the proof of  Theorem~\ref{thm:convergence:intro} we first bound the local gradient $\nabla_r(f_c, t, k)$. We consider the pseudo-network and bound 
\begin{align*}
\mathcal{L}(g_{U(t)}, S(t)) - \mathcal{L}(g_{U^*}, S(t))\leq \alpha(t) + \beta(t) + \gamma(t),
\end{align*}
where 
\begin{align*}
\alpha(t) := & ~ \inner{\wt{\nabla}(f, t)}{ U(t) - U^*}, \\
\beta(t) := & ~ \|\nabla(f,t) - \wt{\nabla}(f,t)\|_{2,1} \cdot \| U(t) - U^* \|_{2,\infty} \\ 
\gamma(t):= & ~\|\nabla(g,t) - \nabla(f,t)\|_{2,1} \cdot \| U(t) - U^* \|_{2,\infty}.
\end{align*}
In bounding $\alpha(t)$, we unfold $\|U(t+1)-U^*\|_F^2$ and have
\begin{align*}
    \alpha(t) = \frac{\eta_{\glo}}{2} \| \Delta U(t) \|_F^2 + \frac{1}{2\eta_{\glo}} \cdot ( \| U(t) - U^* \|_F^2 - \| U(t+1) - U^* \|_F^2 ).
\end{align*}
We bound $\|\Delta U(t)\|_F^2\leq \eta_{\loc}K\cdot o(m)$. By doing summation over $t$, we have
\begin{align*}
    \sum_{t=1}^T \alpha(t)
    = & ~ \frac{\eta_{\glo}}{2} \sum_{t=1}^T \| \Delta U(t) \|_F^2 + \frac{1}{2\eta_{\glo}} \cdot \sum_{t=1}^T ( \| U(t) - U^* \|_F^2 - \| U(t+1) - U^* \|_F^2 ) \\
    \leq & ~ \frac{\eta_{\glo}}{2} \sum_{t=1}^T\| \Delta U(t) \|_F^2 + \frac{1}{2\eta_{\glo}} \cdot   \| U(1) - U^* \|_F^2  \\
    \lesssim & ~ \eta_{\glo}\eta_{\loc} T K \cdot o(m) + \frac{1}{\eta_{\glo}} m D_{U^*}^2
\end{align*}

In bounding $\beta(t)$, we apply Lemma~\ref{lemma:delta_gradient:intro} and have
\begin{align*}
\beta(t) 
= & ~  \|\nabla(f,t) - \wt{\nabla}(f,t)\|_{2,1} \cdot \| U(t) - U^* \|_{2,\infty} \\
\lesssim & ~ o(m) \cdot \| U(t) - U^* \|_{2,\infty} \\
\lesssim & ~ o(m) \cdot (\|U(t)- \wt{U} \|_{2,\infty} + D_{U^*}).
\end{align*}
where $D_{U^*} := \|U^* - \wt{U}\|_{2,\infty} \leq R/m^{c_1}$. As for the first term, we bound
\begin{align*}
    \|U(t)-\wt{U}\|_{2,\infty} \leq & ~ \eta_{\glo}\sum_{\tau=1}^{t}\|\Delta U(\tau)\|_{2,\infty} \\
    = & ~ \eta_{\glo}\sum_{\tau=1}^{t}\|\frac{\eta_{\loc}}{N}\sum_{c=1}^N\sum_{k=0}^{K-1}\nabla(f_c,t,k)\|_{2,\infty} \\
    \leq & ~ \frac{\eta_{\glo}\eta_{\loc}}{N}\sum_{\tau=1}^{t}\sum_{c=1}^N\sum_{k=0}^{K-1}\| \nabla(f_c,t,k)\|_{2,\infty} \\
    \leq & ~ \eta_{\glo}\eta_{\loc}t K m^{-1/3}
\end{align*}
and have $\beta(t)\lesssim \eta_{\glo}\eta_{\loc}t K \cdot o(m) + o(m)\cdot D_{U^*}$, then we do summation and obtain
\begin{align*}
    \sum_{t=1}^T\beta(t) \lesssim \eta_{\glo} \eta_{\loc} T^2 K \cdot o(m) + o(m)\cdot T D_{U^*}.
\end{align*}

In bounding $\gamma(t)$, we apply Lemma~\ref{lemma:coupling:intro} and have
\begin{align*}
\gamma(t) 
& = ~ \|\nabla(g,t) - \nabla(f,t)\|_{2,1} \cdot \| U(t) - U^* \|_{2,\infty} \\
& \lesssim ~ N J \cdot o(m) \cdot (\|U(t)- \wt{U} \|_{2,\infty} + D_{U^*}). 
\end{align*}

Then we do summation over $t$ and have
\begin{align*}
    \sum_{t=1}^T\gamma(t)   
     \lesssim & ~ \eta_{\glo}\eta_{\loc}T^2 K N J \cdot o(m) + T N J \cdot o(m) D_{U^*}
\end{align*}
Putting it together with our choice of our all parameters (i.e. $\eta_{\loc}, \eta_{\glo}, R, K, T, m$), we obtain  
\begin{align*}
  \frac{1}{T}\sum_{\tau=1}^T \mathcal{L}(g_{U(\tau)}, S(\tau)) - \frac{1}{T} \sum_{\tau=1}^T  \mathcal{L}(g_{U^*}, S(\tau)) \leq \frac{1}{T}(\sum_{\tau=1}^T\alpha(\tau)+\sum_{\tau=1}^T\beta(\tau)+\sum_{t=1}^T\gamma(\tau))
  \leq O(\epsilon).
\end{align*}

From Theorem~\ref{thm:real_approximates_pseudo} in appendix, we have $\sup_{x\in \mathcal{X}} |f_U(x)-g_U(x)| \leq O(\epsilon)$
and thus,
\begin{align}
    \label{ineq:1}
   \frac{1}{T} \sum_{t=1}^T \mathcal{L}(f_{U(t)}, \mathcal{S}(t)) - \frac{1}{T} \sum_{t=1}^T \mathcal{L}(f_{U^*}, \mathcal{S}(t)) &\leq O(\epsilon).
\end{align}
From the definition of $\mathcal{A}^*$ we have $\mathcal{L}(f_{U^*}, S(t))\leq \mathcal{L}_{\mathcal{A}^*}(f_{U^*})$.
From the definition of loss we have $\mathcal{L}(f_{U(t)}, \mathcal{S}(t)) = \mathcal{L}_{\mathcal{A}}(f_{U(t)})$.
Moreover, since Eq.~\eqref{ineq:1} holds for all $\epsilon>0$, we can replace $O(\epsilon)$ with $\epsilon$. Thus we prove that for all $\epsilon >0$, 
\begin{align*}
    \frac{1}{T}\sum_{t=1}^T\mathcal{L}_{\mathcal{A}}(f_{U(t)}) \leq \mathcal{L}_{\mathcal{A}^*}(f_{U^*})+\epsilon.
\end{align*}

\paragraph{Combining the results}
From Theorem~\ref{thm:existence:intro} we obtain $U^*$ that is close to $U(0)$ and  makes $\mathcal{L}_{\mathcal{A}^*}(f_{U^*})$ close to zero, from Theorem~\ref{thm:convergence:intro} we have that the average of $\mathcal{L}_{\mathcal{A}}(f_{U(t)})$ is dominated by $\mathcal{L}_{\mathcal{A}^*}(f_{U^*})$. By aggregating these two results, we prove that the minimal of $\mathcal{L}_{\mathcal{A}}(f_{U(t)})$ is $\epsilon$ small and finish the proof of our main Theorem~\ref{thm:main}.

\section{Numerical Results}
\label{sec:numerical_result}

\begin{figure}[h]
    \centering
    \includegraphics[width=0.98\textwidth]{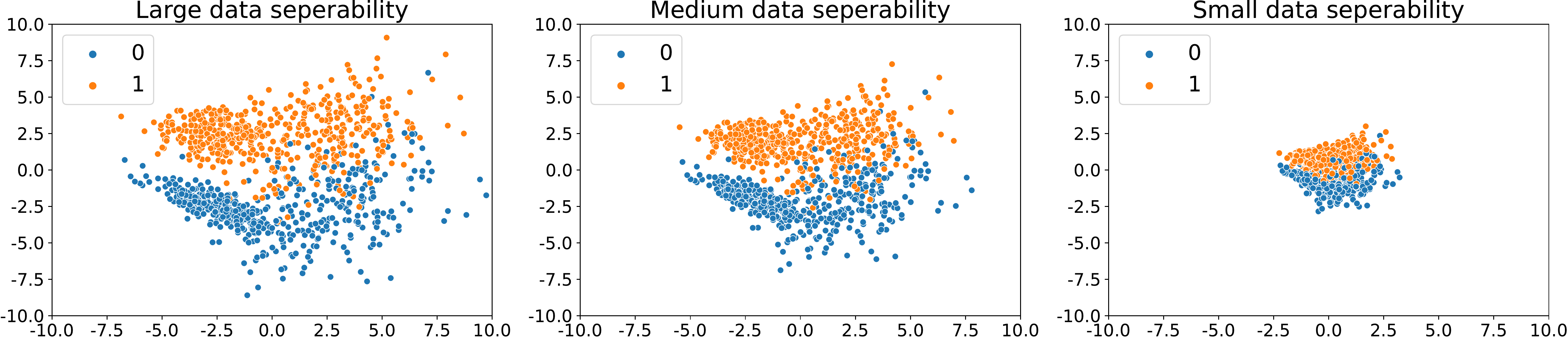}
    \caption{Simulated data with different levels of data separability in numerical experiment.}
    \label{fig:data}
\end{figure}

In this section, we examine our theoretical results (Theorem~\ref{thm:main}) on data separability, $\gamma$ (Definition~\ref{def:sep}), a standard assumption is an over-parameterized neural network convergence analysis. We simulate synthetic data with different levels of data separability as shown in Fig.~\ref{fig:data}. Specifically, each data point contains two dimensions. Each class of data is generated from two Gaussian distributions (std=1) with different means. Each class consists of two Gaussian clusters where the intra-class cluster centroids are closer than the inter-class distances. We perform binary classification tasks on the simulated datasets using multi-layer perceptrons MLP with one hidden layer with 128 neurons. To increase learning difficulty, 5\% of labels are randomly flipped. For each class, we simulated 400 data points as training sets and 100 data points as a testing set. The training data is even divided into four parts to simulate four clients. To simulate different levels of separability, we expand/shrink data features by (2.5, 1.5, 0.85) to construct (large, medium, small) data separability. Note that the whole dataset is not normalized before feeding into the classifier.

\begin{figure*}[h]
    \captionsetup[subfigure]{justification=centering}
    \centering
    \subfloat[FAL ]{\includegraphics[width=0.45\linewidth]{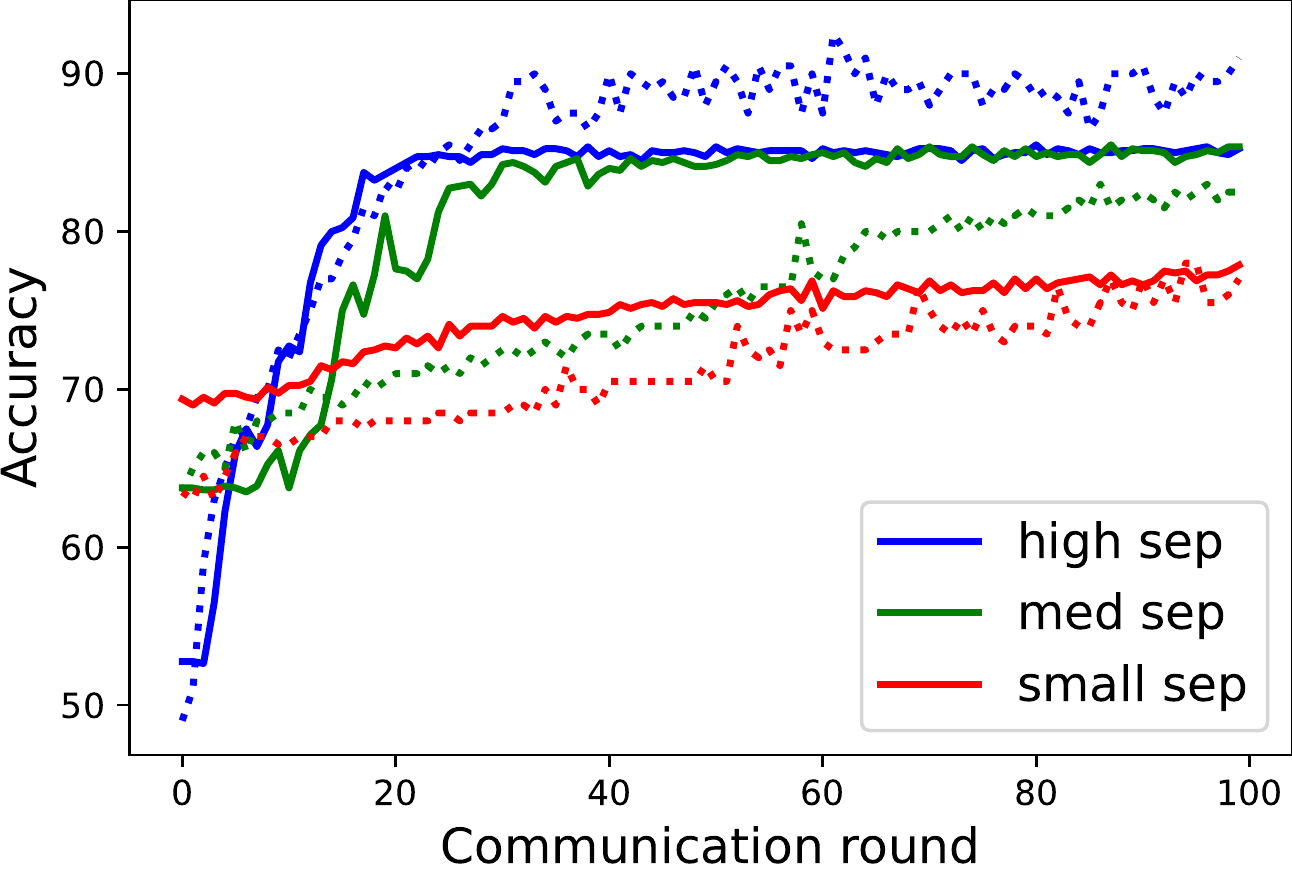}}
    ~
    \subfloat[FedAvg]{\includegraphics[width=0.45\linewidth]{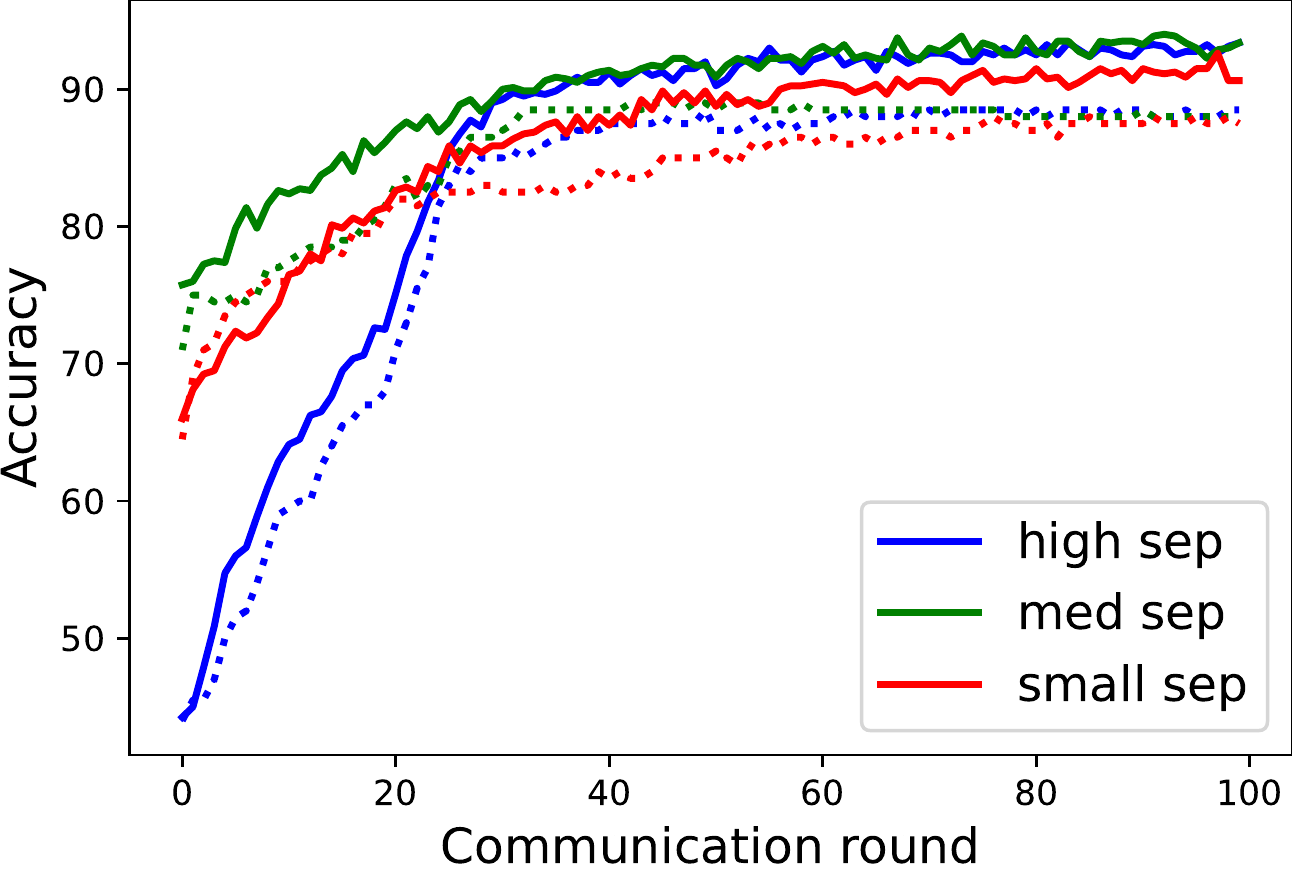}} 
    \caption{Training and testing curve on datasets with different levels of data separability. Solid lines present training curves and dash lines present testing curves.}
    \label{fig:curve}
\end{figure*}

We deploy PGD~\cite{mmstv18} to generate adversarial examples during FAL training with the box of radius $\rho=0.0314$, each perturbation step of 7, and step length of 0.00784. Model aggregation follows FedAvg~\cite{mmr+17} after each local update. The batch size is 50, and  SGD optimizer is used. We depict the training and testing accuracy curves in Fig.~\ref{fig:curve}(a), where solid lines strand for training and dash line stand for testing. The total communication round for is 100, and we observe training convergence for high (blue) and medium (green) separability datasets with learning rate 1e-5. However, a low separability dataset requires a smaller learning rate (i.e., 5e-6) to avoid divergence. From Theorem~\ref{thm:main}, it is easy to see a larger data separability bound $\gamma$ results in a smaller $R$, and we can choose a larger learning rate to achieve convergence. Hence, the selection of learning rate for small separability is consistent with the constraint of learning rate $\eta_{\rm global}$ implied in Theorem~\ref{thm:main}. We empirically observe
results that a dataset with larger data separability $\gamma$ converges faster with the flexibility of choosing a large learning rate,
which is affirmative of our theoretical results that convergence round $T\geq\poly(R/\epsilon)$ has a larger lower bound with a smaller $\gamma$, where $R=\poly(( NJ/\epsilon )^{1/\gamma})
$. In addition, we compare with the accuracy curves obtained by using FedAvg~\cite{mmr+17}. As shown Fig.~\ref{fig:curve}(b), all the datasets converge at around round 40. Therefore, we notice that the same data separability scales have larger affect in FAL training.

\section{Conclusion}
\label{sec:conclusion}
We have studied the convergence of a general format of adopting adversarial training in FL setting to improve FL training robustness. We propose the general framework, FAL, which deploys adversarial samples generation-based adversarial training method on the client-side and then aggregate local model using FedAvg~\cite{mmr+17}. In FAL, each client is trained via min-max optimization with inner loop adversarial generation and outer loop loss minimization. As far as we know, we are the first to detail the proof of theoretical convergence guarantee for over-parameterized ReLU network on the presented FAL strategy, using gradient descent. Unlike the convergence of adversarial training in classical settings, we consider the updates on both local client and global server sides. Our result indicates that we can control learning rates $\eta_{\loc}$ and $\eta_{\glo}$ according to the local update steps $K$ and global communication round $T$ to make the minimal loss close to zero. The technical challenges lie in the multiple local update steps and heterogeneous data, leading to the difficulties of convergence. Under ReLU Lipschitz and over-prameterization assumptions, we use gradient coupling methods twice. Together, we show the model updates of each global updating bounded in our federated adversarial learning. Note that we do not require IID assumptions for data distribution. In sum, the proposed FAL formulation and analysis framework can well handle the multi-local updates and non-IID data in FL. Moreover, our framework can be generalized to other FL aggregation methods, such as sketching and selective aggregation.

\appendix
\ifdefined\isarxiv

\else
\newpage
{
}
\newpage
\fi
\newpage
\paragraph{Roadmap} The appendix is organized as follows. We introduce the probability tools to be used in our proof in Section~\ref{sec:probability_tools}. In addition, we introduce the preliminaries in Section~\ref{sec:pre}. We present the proof overview in Section~\ref{app:sec:overview} and additional remarks used in the proof sketch in Section~\ref{sec:D}. We show the detailed proof for the convergence in Section~\ref{sec:E} and the detailed proof of existence in Section~\ref{sec:F} correspondingly. 

\section{Probability Tools}\label{sec:probability_tools}
We introduce the probability tools that will be used in our proof. First we present two lemmas about random variable's tail bound in
Lemma \ref{lem:chernoff} and \ref{lem:bernstein}:

\begin{lemma}[Chernoff bound \cite{c52}]\label{lem:chernoff}
Let $x = \sum_{i=1}^n x_i$, where $x_i=1$ with probability $p_i$ and $x_i = 0$ with probability $1-p_i$, and all $x_i$ are independent. Let $\mu = \E[x] = \sum_{i=1}^n p_i$. Then \\
1. $ \Pr[ x \geq (1+\delta) \mu ] \leq \exp ( - \delta^2 \mu / 3 ) $, $\forall \delta > 0$ ; \\
2. $ \Pr[ x \leq (1-\delta) \mu ] \leq \exp ( - \delta^2 \mu / 2 ) $, $\forall 0 < \delta < 1$. 
\end{lemma}

\begin{lemma}[Bernstein inequality \cite{b24}]\label{lem:bernstein}
Let $Y_1, \cdots,Y_n$ be independent zero-mean random variables. Suppose that for $i \in [n], |Y_i| \leq M$ almost surely. Then for all $t > 0$, we have
\begin{align*}
\Pr \left[ \sum_{i=1}^n Y_i > t \right] \leq \exp \left( - \frac{ t^2/2 }{ \sum_{i=1}^n \E[Y_i^2]  + M t /3 } \right).
\end{align*}
\end{lemma}

Next, we introduce Lemma \ref{lem:anti_gaussian} about CDF of Gaussian distributions:
\begin{lemma}\label{lem:anti_gaussian}
Let $Z \sim {\N}(0,\sigma^2)$ denotes a Gaussian random variable,
then we have 
\begin{align*}
    \Pr[|Z|\leq t]\in \left( \frac 2 3\frac t \sigma, \frac 4 5\frac t \sigma \right).
\end{align*}
\end{lemma}

Finally, we introduce Claim~\ref{cl:anti-concentration} about elementary anti-concentration property of Gaussian distribution.
\begin{claim}\label{cl:anti-concentration}
    Let $z \sim \mathcal{N}(0, I_d)$ and $u \sim \mathcal{N}(0,1)$ are independent Gaussian random variables. Then for all $t\geq0$ and $x\in\R^d$ that satisfies $\|x\|_2 = 1$, we have
    \begin{align*}
    \Pr[|\langle x, z \rangle + v | \leq t  ] = O( t ) .
    \end{align*}
\end{claim}

\newpage
\section{Preliminaries}
\label{sec:pre}
\subsection{Notations}
For a vector $x$, we use $\| x\|_p$ to denote its $\ell_p$ norm, in this paper we mainly consider the situation when $p=1,2,$ or $\infty$.

For a matrix $U \in \R^{d \times m}$, we use $U^\top$ to denote its transpose and use $\tr[U]$ to denote its trace. We use $\|U\|_1$ to denote its entry-wise $\ell_1$ norm. We use $\|U\|_2$ to denote its spectral norm. We use $\|U\|_F$ to denote its Frobenius norm. For $j \in [m]$, we let $U_j \in \R^d$ be the $j$-th column of $U$. We let $\| U \|_{2,1}$ denotes $\sum_{j=1}^m \|U_j \|_2$. We let $\| U \|_{2,\infty}$ denotes $\max_{j \in [m]} \| U_j \|_2$. For two matrices $X$ and $Y$, we denote their Euclidean inner product as $\inner{X}{Y}:=\tr[X^\top Y]$.

We denote Gaussian distribution with mean $\mu$ and covariance $\Sigma$ as ${\cal N}(\mu,\Sigma)$. We use $\sigma(\cdot)$ to denote the ReLU function, and use $\ind\{A\}$ to denote the indicator function of $A$.

\subsection{Two layer neural network and initialization}
In this paper, we focus on a two-layer neural network that has $m$ neurons in the hidden layer, where each neuron is a ReLU activation function. We define the global network as  
\begin{align}\label{eq:def_f_U}
f_{U}(x) := \sum_{r=1}^m a_r  \cdot \sigma ( \langle U_{r},x \rangle+b_r )
\end{align}
and for $c\in [N]$, we define the local network of client $c$ as
\begin{align}\label{eq:def_f_W}
    f_{W_c}(x) := \sum_{r=1}^m a_r  \cdot \sigma ( \langle W_{c,r},x \rangle+b_r ).
\end{align}
Here $U = (U_{1},U_2,\dots, U_{m})\in \mathbb{R}^{d\times m}$ is the global hidden weight matrix, $W_c = (W_{c,1},\dots, W_{c,m})\in \mathbb{R}^{d\times m}$ is the local hidden weight matrix of client $c$, and $a=(a_1,a_2,\dots,a_m)\in \mathbb{R}^m$ denotes the output weight, $b=(b_1,b_2,\dots,b_m)\in \mathbb{R}^m$ denotes
the bias. 
During the process of federated adversarial learning, for convenience we keep $a$ and $b$ equal to their initialized values and only update $U$ and $W$, so we can write the global network as $f_U(x)$ and the local network as $f_{W_c}(x)$. For the situation we don't care about the weight matrix, we write $f(x)$ or $f_c(x)$ for short.
Next, we make some standard assumptions regarding our training set.

\begin{definition}[Dataset]
There are $N$ clients and $n=N J$ data in total.\footnote{For simplicity, we assume that all clients have same number of training data. Our result can be generalized to the setting where each client has a different number of data as the future work.} Let $\mathcal{S} = \cup_{c \in [N] } \mathcal{S}_c$ where  $\mathcal{S}_c= \{(x_{c,1},y_{c,1}),...,(x_{c,J},y_{c,J}) \} \subseteq \mathbb{R}^d\times \mathbb{R}$ denotes the $J$ training data of client $c$. Without loss of generality, we assume that $\|x_{c,j}\|_2=1$ holds for all $c\in[N], j\in[J]$, and the last coordinate of each point equals to $1/2$, so we consider ${\cal X} : = \{ x \in \R^d : \| x \|_2 =1 ,\ x_{d} = 1/2 \}$. For simplicity, we assume that $|y_{c,j}|\leq 1$ holds for all $c\in[N]$ and $j\in[J]$.\footnote{Our assumptions on data points are reasonable since we can do scale-up. In addition, $l_2$ norm normalization is a typical technique in experiments. Same assumptions also appears in many previous theoretical works like \cite{adhlw19,all19,als19a}.}
\end{definition}

We now define the initialization for the neural networks.
\begin{definition}[Initialization]\label{def:initialize:appendix}
The initialization of $a \in \R^m, U \in \R^{d \times m}, b \in \R^m$ is $a(0) \in \R^m, U(0) \in \R^{d \times m}, b(0) \in \R^m$. The initialization of client c's local weight matrix $W_c$ is $W_c(0,0)=U(0)$. Here the second term in $W_c$ denotes iteration of local steps.
\begin{itemize}
    \item For each $r\in[m]$, $a_r(0)$ are i.i.d. sampled from $ [- 1/ m^{1/3}, +1/m^{1/3}]$ uniformly.
    \item For each $i\in[d], r\in[m]$, $U_{i,r}(0)$ and $b_r(0)$ are i.i.d. random Gaussians sampled from $\mathcal{N}(0, 1/m )$. Here $U_{i,r}$ means the $(i,r)$-entry of $U$.
\end{itemize}
For each global iteration $t\in[T]$,
\begin{itemize}
    \item For each $c\in[N]$, the initial value of client c's local weight matrix $W_c$ is $W_c(t,0)=U(t)$.
\end{itemize}
\end{definition}

\subsection{Adversary and Well-separated training sets}
We first formulate the adversary as a mapping.
\begin{definition}[$\rho$-Bounded adversary]
Let $\mathcal{F}$ denote the function class. An adversary is a mapping $\mathcal{A}: \mathcal{F} \times \mathcal{X} \times \R \rightarrow \mathcal{X}$ which denotes the adversarial perturbation. For $\rho>0$, we define the $\ell_2$ ball as ${\cal B}_2(x,\rho) := \{ \wt{ x}\in \R^d : \| \wt{x} - x\|_2 \leq \rho \}\cap \cal{X}$, we say an adversary $\mathcal{A}$ is \textbf{$\rho$-bounded} if it satisfies
\begin{align*}
    \mathcal{A}(f, x, y) \in \mathcal{B}_2(x,\rho).
\end{align*}
Moreover, given $\rho >0$, we denote the \textbf{worst-case} adversary as $\mathcal{A}^{*} := \argmax_{\tilde{x}\in\mathcal{B}_2(x,\rho)} \ell(f(\tilde{x}), y)$, where $\ell$ is defined in Definition~\ref{def:loss_appendix}.
\end{definition}

In the over-parameterized regime, it is a standard assumption that the training set is well-separated. Since we deal with adversarial perturbations, we require the following $\gamma$-separability, which is a bit stronger.

\begin{definition}[$\gamma$-separability]
Let $\gamma \in (0,1/2), \delta \in (0,1/2), \rho \in (0,1/2)$ denote three parameters such that $\gamma \leq \delta \cdot ( \delta - 2 \rho )$.
We say our training set $\mathcal{S} = \cup_{c \in [N] } \mathcal{S}_c = \cup_{c \in [N], j\in[J] }\{ (x_{c,j}, y_{c,j})\} \subset \R^d \times \R$ is \textbf{globally $\gamma$-separable} w.r.t a $\rho$-bounded adversary, if
\begin{align*}
\min_{c_1 \neq c_2, j_1 \neq j_2} \| x_{c_1,j_1} - x_{c_2,j_2} \|_2 \geq \delta.
\end{align*}
\end{definition}
It is noteworthy that our problem setup does not need the assumption on independent and identically distribution (IID) on data, thus such a formation can be applied to unique challenge of the non-IID setting in FL.

\subsection{Robust loss function}
We define the following Lipschitz convex loss function that will be used.
\begin{definition}[Lipschitz convex loss]\label{def:loss_appendix}
A loss function $\ell: \R\times \R \rightarrow \R$ is said to be a \textbf{Lipschitz convex loss}, if it satisfies the following four properties:
\begin{itemize}
    \item non-negative;
    \item convex in the first input of $\ell$;
    \item $1-$Lipshcitz, which means $\| \ell(x_1, y_1) -\ell(x_2,y_2) \|_2 \leq \|(x_1,y_1)-(x_2,y_2)\|_2$;
    \item $\ell(y, y) = 0$ for all $ y\in\R$.
\end{itemize}
\end{definition}
In this paper we assume $\ell$ is a Lipschitz convex loss. Next, we define our robust loss function of a neural network, which is based on the adversarial examples generated by a $\rho$-bounded adversary $\mathcal{A}$. 

\begin{definition}[Training loss]
Given a client's training set $\mathcal{S}_c = \{ (x_{c,j}, y_{c,j})\}_{j=1}^J \subset \R^d \times \R$ of $J$ samples. 
Let $f_c : \R^d \rightarrow \R$ be a net. We define loss to be $\mathcal{L}(f_c, \mathcal{S}_c) := \frac{1}{J} \sum_{j=1}^J \ell\left(f_c(x_{c,j}), y_{c,j}\right)$. Given $\mathcal{S} = \cup_{c \in [N] } \mathcal{S}_c$, the global loss is defined as 
\begin{align*} 
\mathcal{L}(f_U,\mathcal{S}) := \frac{1}{NJ}\sum_{c=1}^N\sum_{j=1}^J \ell(f_U(x_{c,j}),y_{c,j}).
\end{align*}
Given an adversary $\mathcal{A}$ that is $\rho$-bounded, we define the global loss with respect to $\mathcal{A}$ as
\begin{align*}
    {\cal L}_\mathcal{A}(f_{U}) 
    := & ~ \frac{1}{NJ}\sum_{c=1}^N\sum_{j=1}^J \ell (f_U(\mathcal{A}(f_c, x_{c,j},y_{c,j})), y_{c,j}) \\
    = & ~ \frac{1}{NJ}\sum_{c=1}^N\sum_{j=1}^J \ell (f_U(\tilde{x}_{c,j}), y_{c,j})
\end{align*}
and also define the global robust loss (in terms of \textbf{worst-case}) as
\begin{align*}
    {\cal L}_{\mathcal{A}^*}(f_{U}) := & ~ \frac{1}{NJ}\sum_{c=1}^N\sum_{j=1}^J \ell (f_U(\mathcal{A}^*(f_c, x_{c,j},y_{c,j})), y_{c,j}) \\
    = & ~ \frac{1}{NJ}\sum_{c=1}^N\sum_{j=1}^J \max_{{x}_{c,j}^*\in\mathcal{B}_2(x_{c,j},\rho)}\ell\left(f_U({x}_{c,j}^*), y_{c,j}\right).
\end{align*}
Moreover, since we deal with pseudo-net which is defined in Definition~\ref{def:pseudo}, we also define the loss of a pseudo-net as $\mathcal{L}(g_c, \mathcal{S}_c) :=\frac{1}{J} \sum_{j=1}^J \ell\left(g_c(x_{c,j}), y_{c,j}\right)$ and $\mathcal{L}(g_{U},\mathcal{S}) :=\frac{1}{NJ}\sum_{c=1}^N\sum_{j=1}^J \ell(g_U(x_{c,j}),y_{c,j})$.
\end{definition}

\subsection{Federated Adversarial Learning algorithm}

Classical adversarial training algorithm can be found in \cite{zpdlsa20}. Different from the classical setting, our federated adversarial learning of a local neural network $f_{W_c}$ against an adversary $\mathcal{A}$ is shown in Algorithm~\ref{alg:fl_adv_train}, where there are two procedures: one is \textsc{ClientUpdate} running on client side and the other is \textsc{ServerExecution} running on server side. These two procedures are iteratively processed through communication iterations. Adversarial training is addressed in procedure \textsc{ClientUpdate}. Hence, there are two loops in \textsc{ClientUpdate} procedure: the outer loop is iteration for local model updating; and the inner loop is iteratively generating adversarial samples by the adversary $\mathcal{A}$. In the outer loop in \textsc{ServerExecution} procedure, the neural network's parameters are updated to reduce its prediction loss on the new adversarial samples. These loops constitute an intertwining dynamics.

\begin{algorithm}[!ht]
    \caption{Federated Adversarial Learning (FAL). Complete and formal version of Algorithm~\ref{alg:fl_adv_train_main:intro}. }
    \label{alg:fl_adv_train}
    
	\begin{algorithmic}[1]
	\State {\color{blue}/*Defining notations and parameters*/}
    \State \hspace{4mm} We use $c$ to denote the client's index
    \State \hspace{4mm} The training set of client $c$ is denoted as $\mathcal{S}_c=\{(x_{c,j},y_{c,j})\}_{j=1}^J$
    \State \hspace{4mm} Let $\mathcal{A}$ be the adversary
    \State \hspace{4mm} We denote local learning rate as $\eta_{\loc}$
    \State \hspace{4mm} We denote global learning rate as $\eta_{\mathrm{global}}$
    \State \hspace{4mm} We denote local updating iterations as $K$
    \State \hspace{4mm} We denote global communication round as $T$
    \State
    \State {\color{blue}/*Initialization*/}
	\State \hspace{4mm} {Initialization $a(0) \in \R^m, U(0) \in \R^{d \times m}, b(0) \in \R^m$}
	\State \hspace{4mm} {For $t = 0 \to T$, we iteratively run \textbf{Procedure A} then \textbf{Procedure B}}
	\State 
	\State {\color{blue}/* Procedure running on client side */ }
	\Procedure{\textbf{A}. ClientUpdate}{$t,c$} 
	            \State $\mathcal{S}_c(t) \leftarrow \emptyset$
	            \State $W_{c}(t, 0)\leftarrow U(t)$ \Comment{Receive global model weights update}
	            \For{$k=0 \to K-1$ }
	                \For{$j = 1 \to J$}
	                    \State $\tilde{x}_{c,j}^{(t)} \leftarrow \mathcal{A}(f_{W_{c}(t,k)}, x_{c,j}, y_{c,j} )$ \Comment{Adversarial examples, $f_{W_c}$ is defined as \eqref{eq:def_f_W}}  
	                    \State $\mathcal{S}_c(t) \leftarrow \mathcal{S}_c(t) \cup (\tilde{x}_{c,j}^{(t)}, y_{c,j})$

	                \EndFor
	                \State $W_{c}{(t,k+1)} \leftarrow W_{c}{(t,k)} - \eta_{\loc} \cdot~\nabla_{W_c} \mathcal{L}(f_{W_{c}(t,k)}, \mathcal{S}_c(t) )$
	                \EndFor
	            \State $\Delta U_c(t) \leftarrow W_{c}(t,K) - U(t)$
	    \State Send $\Delta U_c(t)$ to \textsc{ServerExecution}
	\EndProcedure
	\State
	\State {\color{blue}/*Procedure running on server side*/}
    \Procedure{\textbf{B}. ServerExecution} {$t$}:
                \For{each client $c$ \textbf{in parallel}}
                \State{$\Delta U_c(t) \gets$ \textsc{ClientUpdate}$(c,t)$}
                \Comment{Receive local model weights update}
  	        \State $\Delta U(t) \leftarrow \frac{1}{N} \sum_{c \in [N]} \Delta U_c(t)$  
	        \State $U(t+1) \leftarrow U(t) + \eta_{\mathrm{global}} \cdot \Delta U(t)$
                \Comment{Aggregation on the server side}
                \State{Send $U(t+1)$ to client $c$ for \textsc{ClientUpdate}$(c,t)$}
        \EndFor
        \EndProcedure

	\end{algorithmic}
\end{algorithm}     
\clearpage
\newpage
\section{Proof Overview}\label{app:sec:overview}
In this section we give an overview of our main result's proof. Two theorems to be used are Theorem~\ref{thm:convergence} and Theorem~\ref{thm:existence}.

\subsection{Pseudo-network}\label{sec:c1}

We study the over-parameterized neural nets' well-approximated pseudo-network to learn gradient descent for over-parameterized neural nets whose weights are close to initialization. The introducing of pseudo-network makes the proof more intuitive.
 
To be specific, we give the definition of pseudo-network in Section~\ref{sec:D}, and also state Theorem~\ref{thm:real_approximates_pseudo} which shows the fact that the pseudo-network approximates the real network uniformly well. It can be seen that the notion of pseudo-network is used for several times in our proof.

\subsection{Online gradient descent in federated adversarial learning}\label{sec:c2}

Our federated adversarial learning algorithm is formulated in \emph{online gradient descent} framework: at each local step $k$ on the client side, firstly the adversary generates adversarial samples and computes the loss function $\mathcal{L}\left(f_{W_c(t,k)}, \mathcal{S}_c(t)\right)$, then the local client learner takes  the fresh loss function and update $W_c(t,k+1) = W_c(t,k) - \eta_{\loc} \cdot \nabla_{W_c} \mathcal{L}\left(f_{W_c(t,k)}, \mathcal{S}_c(t)\right)$. We refer our readers to \cite{gcl+19,h16} for more details regarding online learning and online gradient descent.

Compared with the centralized setting, the key difficulties in the convergence analysis of FL are induced by multiple local step updates of the client side and the step updates on both local and global sides. Specifically, local updates are not the standard gradient as the centralized adversarial training when $K \geq 2$. We used $-\Delta U(t)$ in substitution of the real gradient of $U$ to update the value of $U(t)$. This brings in challenges to bound the gradient of the neural networks. Nevertheless, gradient bounding is challenging in adversarial training solely. We use gradient coupling method twice to solve this core problem: firstly we bound the difference between real gradient and FL gradient in Lemma~\ref{lemma:delta_gradient}, then we bound the difference between pseudo gradient and real gradient in Lemma~\ref{lemma:coupling}.
We show the connection of online gradient descent and federated adversarial learning in the proof of Theorem~\ref{thm:convergence}.

\subsection{Existence of robust network near initialization}\label{sec:c3}
In Section~\ref{sec:F} we show that there exists a global network $f_{U^*}$ whose weight is close to the initial value $U(0)$ and makes the worst-case global loss $\mathcal{L}_{\mathcal{A}^*}(f_{U^*})$ sufficiently small. We show that the required width $m$ is $\poly (d, ( NJ / \epsilon )^{1/\gamma} )$. 

Suppose we are given a $\rho$-bounded adversary. For a globally $\gamma$-separable training set, to prove Theorem~\ref{thm:existence}, first we state  Lemma~\ref{thm:robust_fitting} which shows the existence of function $f^*$ that has "low complexity" and satisfies $f^*(\tilde{x}_{c,j}) \approx y_{c,j}$ for all data point $(x_{c,j},y_{c,j})$ and perturbation inputs $\tilde{x}_{c,j}\in \mathcal{B}_2(x_{c,j},\rho)$. 
 
Then, we state Lemma~\ref{thm:pseudo_approximates_polynomial} which shows the existence of a pseudo-network $g_{U^*}$
that approximates $f^*$ well. Finally, by using Theorem~\ref{thm:real_approximates_pseudo} we show that $f_{U^*}$ approximates $g_{U^*}$ well. By combining these results, we we finish the proof of Theorem~\ref{thm:existence}.  
\newpage
\section{Real approximates pseudo}\label{sec:D}

To make additional remark to proof sketch in Section~\ref{sec:sketch_main}, in this section, we  state a tool that will be used in our proof that is related to our definition of pseudo-network. First, we recall the definition of pseudo-network.
\begin{definition}[Pseudo-network]\label{def:pseudo}
Given weights $U \in \R^{d \times m}$, $a \in \R^m$ and $b \in \R^{m}$, the global neural network function $f_U : \R^d \rightarrow \R$ is defined as
 
\begin{align*}
    f_U(x) := \sum_{r=1}^{m}a_r\cdot \sigma(\langle U_r, x \rangle+b_r).
\end{align*}
 
Given this $f_U(x)$, we define the corresponding pseudo-network function $g_U : \R^d \rightarrow \R$ as
 
\begin{align*}
g_{U}(x) := \sum_{r=1}^m a_r \cdot \langle U_r(t) - U_r(0) , x \rangle \cdot \ind \{\langle U_r(0), x \rangle + b_r \geq 0 \}.
\end{align*}
\end{definition}

From the definition we can know that pseudo-network can be seen as a linear approximation of the two layer ReLU network we study near initialization.
Next, we cite a Theorem from \cite{zpdlsa20}, which gives a uniform bound of the difference between a network and its pseudo-network.
\begin{theorem}[Uniform approximation, Theorem 5.1 in \cite{zpdlsa20}]\label{thm:real_approximates_pseudo}
 
Suppose $R \geq 1$ is a constant. Let $\rho: = \exp(-\Omega( m^{1/3}) )$. As long as $m \geq \poly(d)$, with prob.  $1-\rho$, for every $U\in\mathbb{R}^{d\times m}$ satisfying $\|U-U(0)\|_{2,\infty}\leq R /  m^{2/3}$, we have
   $
  \sup_{x\in \mathcal{X}} | f_U(x) - g_U(x) |$ is at most $ O ( R^2 / m^{1/6} )$.

   The randomness is due to initialization.
\end{theorem}
\section{Convergence}\label{sec:E}

\begin{table}[h]\caption{List of theorems and lemmas in Section~\ref{sec:E}. The main result of this section is Theorem~\ref{thm:convergence}. By saying "Statements Used" we mean these statements are used in the proof in the corresponding section. For example, Lemma~\ref{lemma:delta_gradient}, \ref{lemma:coupling} and Theorem~\ref{thm:real_approximates_pseudo} are used in the proof of Theorem~\ref{thm:convergence}.}
    \label{tab:overview}
    \centering

    \begin{tabular}{|l|l|l|l|} \hline
        {\bf Section} & {\bf Statement} & {\bf Comment} &
        {\bf Statements Used} \\ \hline
        
       \ref{sec:E_def} & Definition~\ref{def:gradient} and \ref{def:distance} & Definition & - \\ \hline
        \ref{prf:converge} & Theorem~\ref{thm:convergence} & Convergence result & Lem.~\ref{lemma:delta_gradient}, \ref{lemma:coupling}, Thm.~\ref{thm:real_approximates_pseudo} \\ \hline
        
        \ref{sec:delta_gradient} & Lemma~\ref{lemma:delta_gradient} & Approximates real gradient & - \\ \hline
        
        \ref{sec:coupling} & Lemma~\ref{lemma:coupling} & Approximates pseudo gradient & Claim~\ref{claim:aux-couple} \\ \hline
        
        \ref{sec:aux-couple} & Claim~\ref{claim:aux-couple} & Auxiliary bounding & Claim~\ref{cl:anti-concentration} \\ \hline
    \end{tabular}
    
\end{table}

\subsection{Definitions and notations}\label{sec:E_def}
In Section~\ref{sec:E}, we follow the notations used in Definition~\ref{def:pseudo}. Since we are dealing with pseudo-network, we first introduce some additional definitions and notations regarding gradient.  
\begin{definition}[Gradient]\label{def:gradient}
For a local real network $f_{W_c(t,k)}$, we denote its gradient by 
\begin{align*}
    \nabla (f_c,t,k) := & ~ \nabla_{W_c}\mathcal{L}(f_{W_c(t,k)}, \mathcal{S}_c(t)).
\end{align*}
If the corresponding pseudo-network is $g_{W_c(t,k)}$, then we define the pseudo-network gradient as
\begin{align*}
    \nabla (g_c,t,k) := & ~ \nabla_{W_c}\mathcal{L}(g_{W_c(t,k)}, \mathcal{S}_c(t)).
\end{align*}
\end{definition}

Now we consider the global matrix. For convenience we write $\nabla(f,t) := \nabla_U\mathcal{L}(f_{U(t)}, \mathcal{S}(t))$ and $\nabla(g,t) := \nabla_U\mathcal{L}(g_{U(t)}, \mathcal{S}(t))$. We define the \textbf{FL gradient} as $\wt{\nabla}(f,t) := -\frac{1}{N} \Delta U(t)$.
\begin{definition}[Distance]\label{def:distance}
For $U^* \in\mathbb{R}^{d \times m}$ such that $\|U^*-\tilde{U}\|_{2,\infty}\leq R/m^{3/4}$, we define the following distance for simplicity:
\begin{align*}
D_{\max}:= & ~\max_{ t \in [T] }\|\wt{U} - U(t)\|_{2,\infty} \\
D_{U^{*}}:= & ~ \|\wt{U} - U^{*}\|_{2,\infty}
\end{align*}
\end{definition}

We have $D_{U^*}=O(R/m^{3/4})$ and $\|U(t) - U^* \|_{2,\infty} \leq D_{\text{max}} + D_{U^{*}}$ by using triangle inequality.
\begin{table}[h]\caption{Notations of global model weights in federated learning to be used in this section.}
    \centering
    \begin{tabular}{|l|l|l|}
    \hline
        {\bf Notation} & {\bf Meaning} & {\bf Satisfy} \\ \hline
        
        $U(0)$ or $\tilde{U}$ & Initialization of $U$ & $ W_c(0,0) = U(0)$ \\ \hline
        
        $U(t)$ & The value of $U$ after $t$ iterations & $D_{\text{max}}=\max\|U(t) - \wt{U}\|_{2,\infty}$ \\ \hline
        
        $U^*$ & The value of $U$ after small perturbations from $\tilde{U}$ & $ \|U^*-\tilde{U}\|_{2,\infty}\leq R/m^{3/4}$ \\ \hline
    \end{tabular}
      
    \label{tab:overview}
\end{table}

\subsection{Convergence result}\label{prf:converge}
We are going to prove Theorem~\ref{thm:convergence} in this section.
\begin{theorem}[Convergence, formal version of Theorem~\ref{thm:convergence:intro}]\label{thm:convergence}
Let $R \geq 1$.  Suppose $\epsilon\in(0,1)$. Let $K \geq 1$. Let $T\geq \poly(R/\epsilon)$. There is $M = \poly(n, R, 1/\epsilon )$, such that for every $m\geq M$, with probability  $ \geq 1- \exp ( -\Omega (m^{1/3} ) )$, if we run Algorithm~\ref{alg:fl_adv_train} by setting
\begin{align*}
\eta_{\glo}= 1/ \poly(NJ,R,1/\epsilon)~~~ and ~~~ \eta_{\loc} = 1 / K,  
\end{align*}
then for every $U^*$ such that $\| U^* - U(0) \|_{2,\infty} \leq R/m^{3/4}$,  
the output weights $(U(t))_{t=1}^T$ satisfy
\begin{align*}
    \frac{1}{T}\sum_{t=1}^T \mathcal{L}_{\mathcal{A}}\left(f_{U(t)}\right) \leq\mathcal{L}_{\mathcal{A}^*}\left(f_{U^*}\right) + \epsilon.
\end{align*}
The randomness is from $a(0) \in \R^m$, $U(0) \in \R^{d \times m}$, $b(0) \in \R^m $.
\end{theorem}

\begin{proof}
 
We set our parameters as follows:
\begin{align*}
    M &= \Omega\Big(\max\big\{(NJ)^8, (\frac{R}{\epsilon})^{12}\big\}\Big) \\
    \eta_{\glo} &= O(\frac{\epsilon}{Nm^{1/3}\cdot\poly(R/\epsilon)}) \\
    \eta_{\loc} &= 1 / K
\end{align*}

Since the loss function is $1$-Lipschitz, we first bound the $\ell_2$ norm of real net gradient:
\begin{align}\label{ineq:gradient-size}
    \|\nabla_{r}(f_c,t,k)\|_2 \leq |a_r| \cdot  \Big(\frac{1}{J}\sum_{j=1}^J \sigma' (\inner{W_{c,r}(t,k)}{x_{c,j}}+b_r ) \cdot \|\tilde{x}_{c,j}\|_2 \Big) \leq |a_r| \leq \frac{1}{m^{1/3}}.
\end{align}
 
Now we consider the pseudo-net gradient. The loss $\mathcal{L}(g_{U}, \mathcal{S}(t))$ is convex in $U$ due to the fact that $g$ is linear with $U$. Then we have
\begin{align*}
    & ~ \mathcal{L}(g_{U(t)}, \mathcal{S}(t)) - \mathcal{L}(g_{U^*}, \mathcal{S}(t)) \\
    \leq & ~ \inner{\nabla_U\mathcal{L}(g_{U(t)}, \mathcal{S}(t))}{U(t)- U^*} \\
    = & ~ \inner{\wt{\nabla} (f,t)}{U(t) - U^*} + \inner{\nabla(f,t) - \wt{\nabla}(f,t)}{U(t) - U^*} + \inner{ \nabla(g,t) - \nabla(f,t)}{U(t) - U^*} \\
    \leq & ~ \alpha(t)+\beta(t)+\gamma(t)
\end{align*}
where the last step follows from 
\begin{align*}
\alpha(t) := & ~ \inner{\wt{\nabla}(f, t)}{ U(t) - U^*} , \\
\beta(t) := & ~ \|\nabla(f,t) - \wt{\nabla}(f,t)\|_{2,1} \cdot \| U(t) - U^* \|_{2,\infty}, \\
\gamma(t) := & ~ \|\nabla(g,t) - \nabla(f,t)\|_{2,1} \cdot \| U(t) - U^* \|_{2,\infty}.
\end{align*}
Note that the FL gradient $\wt{\nabla}(f,t) = -\frac{1}{N}\Delta U(t)$ is the direction moved by center, in contrast, $\nabla(f,t)$ is the true gradient of function $f$.
We deal with these three terms separately. As for $\alpha(t)$, we have  
\begin{align*}
\| U(t+1) - U^* \|_F^2
= & ~ \| U(t)+\eta_{\glo} \Delta U(t) - U^* \|_F^2 \\
= & ~ \| U(t) - U^* \|_F^2 - 2N \eta_{\glo} \alpha(t)+\eta_{\glo}^2  \|\Delta U(t) \|_{F}^2
\end{align*}
and by rearranging the equation we get
\begin{align*}
\alpha(t) = \frac{\eta_{\glo}}{2N} \| \Delta U(t) \|_F^2 + \frac{1}{2N \eta_{\glo}} \cdot ( \| U(t) - U^* \|_F^2 - \| U(t+1) - U^* \|_F^2 ).
\end{align*}
Next, we need to upper bound $\| \Delta U(t) \|_F^2$, 
\begin{align}\label{eq:bound_Delta_U(t)}
  \| \Delta U(t) \|_F^2
  & = \| \frac{\eta_{\loc}}{N}\sum_{c=1}^N \sum_{k=0}^{K-1} \nabla (f_c,t,k) \|_F^2 \notag \\
  & \leq \frac{\eta_{\loc}}{N}\sum_{c=1}^N \sum_{k=0}^{K-1}\sum_{r=1}^m \|\nabla_r(f_c,t,k)\|_2^2 \notag \\
  & = \eta_{\loc}Km^{1/3} \notag \\
  & = m^{1/3}.
\end{align}
where the last step follows from $K\eta_{\loc} = 1$. Then we do summation over $t$ and have 
\begin{align*}
    \sum_{t=1}^T \alpha(t) = & ~ \frac{\eta_{\glo}}{2N} \sum_{t=1}^T \| \Delta U(t) \|_F^2 + \frac{1}{2N\eta_{\glo}} \cdot \sum_{t=1}^T ( \| U(t) - U^* \|_F^2 - \| U(t+1) - U^* \|_F^2 ) \\
    = & ~ \frac{\eta_{\glo}}{2N} \sum_{t=1}^T\| \Delta U(t) \|_F^2 + \frac{1}{2N\eta_{\glo}} \cdot  ( \| U(1) - U^* \|_F^2 - \| U(T+1) - U^* \|_F^2 ) \\
    \leq & ~ \frac{\eta_{\glo}}{2N} \sum_{t=1}^T\| \Delta U(t) \|_F^2 + \frac{1}{2N\eta_{\glo}} \cdot \| U(1) - U^* \|_F^2  \\
    \lesssim & ~ \frac{\eta_{\glo}}{N} T m^{1/3} + \frac{1}{N\eta_{\glo}} m D_{U^*}^2
\end{align*}

where the last step follows from Eq.~\eqref{eq:bound_Delta_U(t)} and $\|\wt{U} - U^*\|_F^2\leq m\cdot \|\wt{U} - U^*\|_{2,\infty}^2=m D_{U^*}^2$.

As for $\beta(t)$, we apply Lemma~\ref{lemma:delta_gradient} and also triangle inequality and have
\begin{align*}
\beta(t) 
= & ~  \|\nabla(f,t) - \wt{\nabla}(f,t)\|_{2,1} \cdot \| U(t) - U^* \|_{2,\infty} \\
\lesssim & ~ m^{2/3} \cdot \| U(t) - U^* \|_{2,\infty} \\
\lesssim & ~ m^{2/3} \cdot (\|U(t)- \wt{U} \|_{2,\infty} + D_{U^*}).
\end{align*}
By using Eq.~\eqref{ineq:gradient-size} we bound the size of $\|U(t)-\wt{U}\|_{2,\infty}$:
\begin{align*}
    \|U(t)-\wt{U}\|_{2,\infty} & \leq \eta_{\glo}\sum_{\tau=1}^{t}\|\Delta U(\tau)\|_{2,\infty} \\
    & = \eta_{\glo}\sum_{\tau=1}^{t}\|\frac{\eta_{\loc}}{N}\sum_{c=1}^N\sum_{k=0}^{K-1}\nabla(f_c,t,k)\|_{2,\infty} \\
    & \leq \frac{\eta_{\glo}\eta_{\loc}}{N}\sum_{\tau=1}^{t}\sum_{c=1}^N\sum_{k=0}^{K-1}\| \nabla(f_c,t,k)\|_{2,\infty} \\
    & \leq \eta_{\glo}\eta_{\loc}t K m^{-1/3}
\end{align*}
and have
\begin{align*}
    \beta(t)\lesssim \eta_{\glo}\eta_{\loc}t K m^{1/3} + m^{2/3}D_{U^*}.
\end{align*}
Then we do summation over $t$ and have
\begin{align*}
    \sum_{t=1}^T\beta(t) & \lesssim \sum_{t=1}^T(\eta_{\glo}\eta_{\loc}t K m^{1/3} + m^{2/3}D_{U^*}) \\
    & \lesssim \eta_{\glo}\eta_{\loc}T^2 K m^{1/3} + m^{2/3} T D_{U^*} \\
    & \lesssim \eta_{\glo}T^2 m^{1/3} + m^{2/3} T D_{U^*}.
\end{align*}

As for $\gamma(t)$, we apply Lemma~\ref{lemma:coupling}  
and have
\begin{align*}
\gamma(t) 
= & ~  \|\nabla(g,t) - \nabla(f,t)\|_{2,1} \cdot \| U(t) - U^* \|_{2,\infty} \\
\lesssim & ~ N J m^{13/24} \cdot (\|U(t)- \wt{U} \|_{2,\infty} + D_{U^*}). 
\end{align*}

Since $\|U(t)-\wt{U}\|_{2,\infty} \leq \eta_{\glo}\eta_{\loc}t K m^{-1/3}$, we have
\begin{align*}
    \gamma(t)\lesssim \eta_{\glo}\eta_{\loc}t K N J m^{5/24} + N J m^{13/24} D_{U^*}.
\end{align*}
 
Then we do summation over $t$ and have
\begin{align*}
    \sum_{t=1}^T\gamma(t) & \lesssim \sum_{t=1}^T\big(\eta_{\glo}\eta_{\loc}t K N J m^{5/24} + N J m^{13/24} D_{U^*}\big) \\
    & \lesssim \eta_{\glo}\eta_{\loc}T^2 K N J m^{5/24} + N J m^{13/24}T D_{U^*} \\
    & \lesssim \eta_{\glo}T^2 N J m^{5/24} + N J m^{13/24}T D_{U^*}.
\end{align*}
Next we put it altogether. Note that $D_{U^*}=O (\frac{R}{m^{3/4}} )$, thus we obtain 
\begin{align*}
    & \sum_{t=1}^T\mathcal{L}(g_{U(t)}, \mathcal{S}(t)) - \sum_{t=1}^T\mathcal{L}(g_{U^*}, \mathcal{S}(t)) \\
    \leq & \sum_{t=1}^T\alpha(t) + \sum_{t=1}^T\beta(t) + \sum_{t=1}^T\gamma(t)\\
    \lesssim & ~ \frac{\eta_{\glo}}{N} T m^{1/3} + \frac{1}{N\eta_{\glo}} m D_{U^*}^2 + \eta_{\glo}T^2 m^{1/3} \\ & + m^{2/3} T D_{U^*} + \eta_{\glo}T^2 N J m^{5/24} + N J m^{13/24}T D_{U^*}\\
    \lesssim & ~ \frac{\eta_{\glo}}{N} T m^{1/3} + \frac{1}{N\eta_{\glo}} R^2 m^{-1/2} + \eta_{\glo}T^2 m^{1/3} \\ &+ R T m^{-1/12} + \eta_{\glo}T^2 N J m^{5/24} + N J m^{-5/24} R T.
\end{align*}
We then have
\begin{align}
\label{eq:12}
  & \frac{1}{T}\sum_{\tau =1}^T \mathcal{L}(g_{U(\tau )}, \mathcal{S}(\tau )) - \frac{1}{T} \sum_{\tau =1}^T  \mathcal{L}(g_{U^*}, \mathcal{S}(\tau )) \notag \\
  \lesssim ~ & \frac{\eta_{\glo}}{N} m^{1/3} + \frac{1}{N \eta_{\glo} T} R^2 m^{-1/2} + \eta_{\glo}T m^{1/3} + Rm^{-1/12} \notag \\ &+ \eta_{\glo}T N J m^{5/24} + N J m^{-5/24} R. \notag \\
  \lesssim ~ & \frac{1}{N \eta_{\glo} T} R^2 m^{-1/2} + \eta_{\glo}T m^{1/3} + Rm^{-1/12} + \eta_{\glo}T N J m^{5/24} + N J m^{-5/24} R \\
  \leq ~ & O(\varepsilon). \notag
\end{align}

From Theorem~\ref{thm:real_approximates_pseudo} we know
\begin{align*}
    \sup_{x\in \mathcal{X}} |f_U(x)-g_U(x)| \leq O(R^2/m^{1/6}) = O(\epsilon).
\end{align*}
 
 In addition,
\begin{align}
    \label{ineq:2}
   \frac{1}{T} \sum_{t=1}^T \left(  \mathcal{L}(f_{U(t)}, \mathcal{S}(t)) - \mathcal{L}(f_{U^*}, \mathcal{S}(t)) \right) &\leq  O( \varepsilon )
\end{align}

From the definition of $\mathcal{A}^*$ we have $\mathcal{L}(f_{U^*}, S(t))\leq \mathcal{L}_{\mathcal{A}^*}(f_{U^*})$. From the definition of loss we have $\mathcal{L}(f_{U(t)}, \mathcal{S}(t)) = \mathcal{L}_{\mathcal{A}}(f_{U(t)})$. Moreover, since Eq.~\eqref{ineq:2} holds for all $\epsilon>0$, we can replace $\frac{\epsilon}{c}$ with $\epsilon$. Thus we prove that for $\forall \epsilon >0$, 
\begin{align*}
    \frac{1}{T}\sum_{t=1}^T\mathcal{L}_{\mathcal{A}}(f_{U(t)}) \leq \mathcal{L}_{\mathcal{A}^*}(f_{U^*})+\epsilon.
\end{align*}

\end{proof}

\subsection{Approximates real global gradient}\label{sec:delta_gradient}
We are going to prove Lemma~\ref{lemma:delta_gradient} in this section.

\begin{lemma}[Bounding the difference between real gradient and FL gradient]\label{lemma:delta_gradient}
Let $\rho := \exp(-\Omega(m^{1/3}))$
With probability $\geq 1-\rho$, for iterations $t$ satisfying
 
\begin{align*}
    | U(t) - U(0)\|_{2,\infty} \leq O ( m^{-15/24} ),
\end{align*}
the following holds:
\begin{align*}
  \| \nabla (f, t) - \wt{\nabla}(f,t) \|_{2,1} \leq O(m^{2/3}) .
\end{align*}
The randomness is from $a(\tau) \in \R^m$, $U(\tau) \in \R^{d \times m}$, $b(\tau) \in \R^m $ for $\tau$ at $0$.
\end{lemma}
\begin{proof}
Notice that $\nabla(f,t) = \nabla_U\mathcal{L}(f_{U(t)}, \mathcal{S}(t))$ and
\begin{align*}
    \wt{\nabla}(f,t) = -\frac{1}{N}\Delta U(t) = -\frac{1}{N}\sum_{c=1}^N \Delta U_c(t) =\frac{\eta_{\loc}}{N} \sum_{c=1}^N\sum_{k=0}^{K-1}\nabla(f_c,t,k).
\end{align*}
So we have
\begin{align*}
    \| \nabla (f, t) - \wt{\nabla}(f,t) \|_{2,1} & =\sum_{r=1}^m \|\nabla_r(f, t) - \wt{\nabla}_r(f,t) \|_{2} \\
    & = \frac{1}{N}\sum_{r=1}^m \|N\cdot\nabla_r(f,t) -\eta_{\loc} \sum_{c=1}^N \sum_{k=0}^{K-1}\nabla_r(f_c,t,k)\|_2 \\
    & \leq \frac{\eta_{\loc}}{N}\sum_{r=1}^m\sum_{k=0}^{K-1}\|\frac{N\cdot\nabla_r(f,t)}{K\eta_{\loc}} -\sum_{c=1}^N \nabla_r(f_c,t,k)\|_2 \\
    & = \frac{1}{N K}\sum_{r=1}^m\sum_{k=0}^{K-1}\|N\cdot \nabla_r(f,t) - \sum_{c=1}^N\nabla_r(f_c,t,k)\|_2
\end{align*}
where the last step follows from the assumption that $\eta_{\loc}=\frac{1}{K}$.

As for $\|N\cdot \nabla_r(f,t) - \sum_{c=1}^N \nabla_r(f_c,t,k)\|_2$, we have
\begin{align*}
    & \|N\cdot \nabla_r(f,t) - \sum_{c=1}^N\nabla_r(f_c,t,k)\|_2 \\
    \leq & ~ |a_r|\cdot \Big|\big(\frac{N}{NJ}\sum_{c=1}^N\sum_{j=1}^J \ind\{\inner{U_r(t)}{x_{c,j}}+b_r\geq 0\} \\ &- \frac{1}{J}\sum_{c=1}^N\sum_{j=1}^J \ind\{\inner{W_{c,r}(t,k)}{x_{c,j}}+b_r\geq 0\} \big) \cdot \|x_{c,j}\|_2 \Big| \\ 
    \leq & ~ \frac{1}{m^{1/3}}\cdot\frac{1}{J}\sum_{c=1}^N\sum_{j=1}^J |\ind\{\inner{U_r(t)}{x_{c,j}}+b_r\geq 0\} - \ind\{\inner{W_{c,r}(t,k)}{x_{c,j}}+b_r\geq 0\}| \\
    \leq & ~ \frac{N}{m^{1/3}}.
\end{align*}
Then we do summation and have
\begin{align*}
    \| \nabla (f, t) - \wt{\nabla}(f,t) \|_{2,1} \leq & ~ \frac{1}{N K}\sum_{r=1}^m\sum_{k=0}^{K-1}\|N\cdot \nabla_r(f,t) - \sum_{c=1}^N\nabla_r(f_c,t,k)\|_2 \\
    \leq & ~ \frac{1}{NK}\sum_{r=1}^m\sum_{k=0}^{K-1}\frac{N}{m^{1/3}} \\
    = & ~ m^{2/3}.
\end{align*}
Thus we finish the proof.
\end{proof}

\subsection{Approximates pseudo global gradient}\label{sec:coupling}
We are going to prove Lemma~\ref{lemma:coupling} in this section.
\begin{lemma}[
]
\label{lemma:coupling}
Let $\rho: = \exp(-\Omega(m^{1/3}))$.
With probability   $ \geq 1- \rho$, for iterations $t$ satisfying 
 
\begin{align*}
    \| U(t) - U(0)\|_{2,\infty} \leq O ( m^{-15/24} ),
\end{align*}
the following holds:
\begin{align*}
  \| \nabla (g, t) - \nabla(f,t) \|_{2,1} \leq O(N J m^{13/24} ) .
\end{align*}
The randomness is because $a(\tau) \in \R^m$, $U(\tau) \in \R^{d \times m}$, $b(\tau) \in \R^m $, for $\tau = 0$.
\end{lemma}
\begin{proof}
 
Notice that $\nabla(g,t) = \nabla_U\mathcal{L}(g_{U(t)}, \mathcal{S}(t))$ and $\nabla(f,t)= \nabla_U\mathcal{L}(f_{U(t)}, \mathcal{S}(t))$. By Claim~\ref{claim:aux-couple}, with the given probability we have
\begin{align*}
    \sum_{r=1}^m \ind \{\nabla_{r}(g,t) \not= \nabla_{r}(f,t)\} \leq O(N J m^{7/8} ).
\end{align*}

For indices $r \in [m]$ satisfying $ \nabla_{r}(g,t) \not= \nabla_{r}(f,t)$, the following holds:
\begin{align*}
\| \nabla_{r}(g,t) - \nabla_{r}(f,t) \|_2
= & ~ \|\nabla_{U,r}\mathcal{L}(g_{U(t)}, S(t)) -\nabla_{U,r}\mathcal{L}(f_{U(t)},S(t))\|_2 \\
\leq & ~ |a_r|\cdot \frac{1}{N J} \cdot \sum_{c=1}^N \sum_{j=1}^J \|x_{c,j}\|_2 \cdot \big| \ind\{\inner{\wt{U}_r} {x_{c,j}} + b_r \geq 0 \} \\& - \ind\{\inner{U_r}{x_{c,j}} + b_r\geq 0 \} \big| \\
\leq & ~ \frac{1}{m^{1/3}}\cdot \frac{1}{NJ} \cdot \sum_{c=1}^N \sum_{j=1}^J \big| \ind\{\inner{\wt{U}_r} {x_{c,j}} + b_r \geq 0 \} - \ind\{\inner{U_r}{x_{c,j}} + b_r\geq 0 \} \big| \\
\leq & ~ \frac{1}{m^{1/3}} .
\end{align*}
where the first step is definition, the second step follows that the loss function is $1$-Lipschitz, the third step follows from $|a_r|\leq\frac{1}{m^{1/3}}$ and $\|x_{c,j}\|_2 = 1$, the last step follows from the bound of the indicator function.
Thus, we do the conclusion:
\begin{align*}
    & ~ \| \nabla(g,t) - \nabla(f,t) \|_{2,1} \\
    = & ~ \sum_{r=1}^m \| \nabla_{r} (g,t) - \nabla_r(f,t) \|_2\cdot\ind \{\nabla_{r}(g,t) \not= \nabla_{r}(f,t)\} \\
    \leq & ~ \frac{1}{m^{1/3}}\sum_{r=1}^m\ind \{\nabla_{r}(g,t) \not= \nabla_{r}(f,t)\} \\
    \leq & ~ \frac{1}{m^{1/3}} \cdot O(N J m^{7/8} ) \\
    = & ~ O(N J m^{13/24} )
\end{align*}
and finish the proof.
\end{proof}

\subsection{Bounding auxiliary}\label{sec:aux-couple}
 
\begin{claim}[Bounding auxiliary]\label{claim:aux-couple}
Let $\rho : = \exp(-\Omega(m^{1/3}))$.
With probability   $\geq 1- \rho$ , we have
\begin{align*}
    \sum_{r=1}^m \ind \{\nabla_{r}(g,t) \not= \nabla_{r}(f,t)\} \leq O(N J m^{7/8} ).
\end{align*}
The randomness is from $a(\tau) \in \R^m$, $U(\tau) \in \R^{d \times m}$, $b(\tau) \in \R^m $ for $\tau = 0$.
\end{claim}

\begin{proof}
For $r\in[m]$, let $I_r := \ind \{\nabla_{r}(g,t) \not= \nabla_{r}(f,t)\}$.
By Claim~\ref{cl:anti-concentration} we know that for each $x_{c,j}$ we have  
\begin{align*}
    \Pr [|\inner{ \wt{W}_{c,r} }{x_{c,j}}+b_r|\leq  m^{-15/24}  ]\leq O( m^{-1/8} ).
\end{align*}
By putting a union bound on $c$ and $j$, we get
\begin{align*}
    \Pr \big[\exists c\in[N], j\in[J], ~|\inner{ \wt{W}_{c,r} }{x_{c,j}}+b_r|\leq m^{-15/24} \big] \leq O(N J m^{-1/8} ).
\end{align*}
Since
\begin{align*}
    \Pr [ I_r = 1 ] \leq \Pr [\exists j\in[J], c\in[N],  ~|\inner{ \wt{W}_{c,r} }{x_{c,j}}+b_r|\leq m^{-15/24}  ],
\end{align*}
we have
\begin{align*}
     \Pr [ I_r = 1 ] \leq O (N J m^{-1/8} ).
\end{align*}
By applying concentration inequality on $I_r$ (independent Bernoulli) for $r\in [m]$, we obtain that with prob. 
  
\begin{align*}
    \geq 1-\exp(-\Omega(N J m^{7/8})) > 1-  \rho,
\end{align*}
the following holds:
\begin{align*}
    \sum_{r=1}^m I_r \leq O (N J m^{7/8} ).
\end{align*}
Thus we finish the proof.
 
\end{proof}

\subsection{Further Discussion}
Note that in the proof of Theorem~\ref{thm:convergence} we set the hidden layer's width $m$ to be greater than $O(\epsilon^{-12})$, which seems impractical in reality: if we choose our convergence accuracy to be $10^{-2}$, the width will become $10^{24}$ which is impossible to achieve.

However, we want to claim that the "$-12$" term is not intrinsic in our theorem and proof, and we can actually further improve the lower bound of $m$ to $O((R/\epsilon)^{c_2})$ where $c_2$ is some constant between $-3$ and $-4$. To be specific, we observe from Eq.~\eqref{eq:12} that the "$-12$" term comes from $\frac{2}{3} - \frac{3}{4} = -\frac{1}{12}$, where $\frac{2}{3}$ appears in Lemma~\ref{lemma:delta_gradient} and $\frac{3}{4}$ appears in the assumption that $D_{U^*}\leq R/m^{3/4}$ in Definition~\ref{def:distance}. As for our observations, the $\frac{2}{3}$ term is hard to improve. On the other hand, we can actually adjust the value of $D_{U^*}$ as long as we ensure
\begin{align*}
    D_{U^*} \leq R/ m^{c_3}
\end{align*}
for some constant $c_3\in (0, 1)$. When we let $c_3\to 1$, the final result will achieve
\begin{align*}
    O(({R}/{\epsilon})^3)
\end{align*}
which is much more feasible in reality.

As the first work and the first step towards understanding the convergence of federated adversarial learning, the priority of our work is not achieving the tightest bounds. Instead, our main goal is to show the convergence of a general federated adversarial learning framework. Nevertheless, we will improve the bound in the final version.

\section{Existence}\label{sec:proofs}\label{sec:F}

In this section we prove the existence of $U^*$ that is close to $U(0)$ and makes $\mathcal{L}_{\mathcal{A}^*}(f_{U^*})$ close to zero.
\subsection{Tools from previous work}
In order to prove our existence result, we first state two lemmas that will be used.

\begin{lemma}[Lemma 6.2 from~\cite{zpdlsa20}]\label{thm:robust_fitting}
Suppose that $ \|x_{c_1,j_1}-x_{c_2,j_2}\|_2\geq \delta$ holds for each pair of two different data points $x_{c_1,j_1}, x_{c_2,j_2}$. Let $D = 24 \gamma^{-1} \ln ( 48 NJ / \epsilon )$, then there $\exists$ a polynomial $g : \R \rightarrow \R$ with size of coefficients no bigger than $O(\gamma^{-1}2^{6D})$ and degree no bigger than $D$, that satisfies for all $c_0 \in [N], j_0\in [J]$ and $\tilde{x}_{c_0,j_0}\in \mathcal{B}_2(x_{c_0,j_0},\rho)$,
\begin{align*}
\left|\sum_{c=1}^N\sum_{j=1}^J y_{c,j} \cdot g( \langle x_{c,j} , \tilde{x}_{c_0,j_0} \rangle )-y_{c_0,j_0} \right| \leq \frac{\epsilon}{3}.
\end{align*}
\end{lemma}

We let $f^*(x):=\sum_{c=1}^N\sum_{j=1}^J y_{c,j} \cdot g( \langle x_{c,j} , x \rangle )$ and have $|f^*(\tilde{x}_{c_0,j_0})-y_{c_0,j_0}|\leq \epsilon /3$.

\begin{lemma}[Lemma 6.5 from~\cite{zpdlsa20}]\label{thm:pseudo_approximates_polynomial}
Suppose $\epsilon\in(0,1)$. Suppose
\begin{align*}
    M=\poly (( NJ / \epsilon )^{1/\gamma} , d ) ~~~ \text{and} ~~~ R=\poly ( ( NJ / \epsilon )^{1/\gamma} )
\end{align*}
As long as $m\geq M$, with prob. $ \geq 1- \exp( -\Omega ( \sqrt{m/NJ}))$ , there $ \exists U^* \in \mathbb{R}^{d\times m}$ that satisfies
\begin{align*}
    \| U^*-U(0) \|_{2,\infty} \leq R / m^{2/3} ~~~ \text{and} ~~~
    \sup_{x\in\cal{X}} |g_{U^*}(x) -f^*(x) | \leq \epsilon/3.
    \end{align*}
The randomness is due to $a(\tau) \in \R^m$, $U(\tau) \in \R^{d \times m}$, $b(\tau) \in \R^m $ for $\tau = 0$.
\end{lemma}

\subsection{Existence result}\label{sec:main_result}
We are going to present Theorem~\ref{thm:existence} in this section and present its proofs.
\begin{theorem}[Existence, formal version of Theorem~\ref{thm:existence:intro}]\label{thm:existence}
Suppose that $\epsilon\in(0,1)$. Suppose
 \begin{align*}
 M_0 = \poly (d, ( NJ / \epsilon )^{1/\gamma} )   ~~~ and ~~~   R = \poly( ( NJ / \epsilon )^{1/\gamma} )
 \end{align*}
As long as $m \geq M_0$, then with prob. $\geq 1- \exp ( -\Omega (m^{1/3} )  )$, there exists $U^*\in\mathbb{R}^{d\times m}$ satisfying
\begin{align*}
    \|U^*-U(0)\|_{2,\infty} \leq R / m^{2/3} ~~~ \text{and} ~~~
    \mathcal{L}_{\mathcal{A}^*} ( f_{U^*} ) \leq \epsilon .
    \end{align*}
The randomness comes from $a(\tau) \in \R^m$, $U(\tau) \in \R^{d \times m}$, $b(\tau) \in \R^m $ for $\tau = 0$.
\end{theorem}

\begin{proof}

For convenient, we define
\begin{align*}
    \rho_0 := \exp(-\Omega(\sqrt{m/NJ}))-\exp(-\Omega(m^{1/3})).
\end{align*}

From Lemma \ref{thm:robust_fitting} we obtain the function $f^*$. From Lemma \ref{thm:pseudo_approximates_polynomial} we know the existence of $M_0=\poly(d, (NJ/\epsilon)^{1/\gamma})$ and also $R=\poly((NJ/\epsilon)^{1/\gamma})$.

By combining these two results with Theorem \ref{thm:real_approximates_pseudo}, we have that for all $m\geq \poly(d, (NJ/\epsilon)^{1/\gamma})$, with prob. 
\begin{align*}
   \geq  1- \rho_0,
\end{align*}
there $ \exists U^*\in\mathbb{R}^{d\times m}$ that satisfies $\|U^*-U(0)\|_{2,\infty} \leq R/ m^{2/3}$.

In addition, the following properties:
\begin{itemize}
  \item
  $\max_{x \in {\cal X}} |g_{U^*}(x)-f^*(x)|$ is at most $\epsilon/3$ 
  \item 
  $\max_{x \in {\cal X}} |f_{U^{*}}(x)-g_{U^{*}}(x)|$ is at most $O( R^2/ m^{1/6} )$
\end{itemize}

Consider the loss function. For all $c\in[N]$, $j\in[J]$ and $\tilde{x}_{c,j}\in \mathcal{B}(x_{c,j},\rho)$, we have
\begin{align*}
   \ell(f_{U^*}(\tilde{x}_{c,j}),y_{c,j}) 
   \leq & ~ |f_{U^*}(\tilde{x}_{c,j})-y_{c,j}| \\
   \leq & ~ |f_{U^{*}}(\tilde{x}_{c,j})-g_{U^{*}}
   (\tilde{x}_{c,j})| + |g_{U^{*}}(\tilde{x}_{c,j})-f^*(\tilde{x}_{c,j})| + |f^*(\tilde{x}_{c,j})-y_{c,j}| \\
   \leq & ~ O( R^2 / m^{1/6}) +\frac{\epsilon}{3} +\frac{\epsilon}{3} \\
   \leq & ~ \epsilon,
\end{align*}
Thus, we have that
\begin{align*}
    \mathcal{L}_{\mathcal{A}^*}(f_{U^*}) = \frac{1}{NJ}\sum_{c=1}^N\sum_{j=1}^J \max \ell\left(f_{U^*}({x}_{c,j}^*), y_{c,j}\right)\leq \epsilon.
\end{align*}
Furthermore, since the $m$ we consider satisfies $m\geq\Omega((NJ)^{1/\gamma})$, the holding probability is 
\begin{align*}
    \geq & ~ 1- \rho_0 \\
    =  & ~ 1- \exp(-\Omega(m^{1/3})).
\end{align*}
Thus, it finishes the proof of this theorem.

\end{proof}

\ifdefined\isarxiv
\bibliographystyle{alpha}
\bibliography{ref}
\else
\bibliographystyle{alpha}
\bibliography{ref}

\newcommand{\etalchar}[1]{$^{#1}$}
\begin{thebibliography}{ADH{\etalchar{+}}19b}

\bibitem[ACW18]{acw18}
Anish Athalye, Nicholas Carlini, and David Wagner.
\newblock Obfuscated gradients give a false sense of security: Circumventing
  defenses to adversarial examples.
\newblock {\em arXiv preprint arXiv:1802.00420}, 2018.

\bibitem[ADH{\etalchar{+}}19a]{adhlsw19}
Sanjeev Arora, Simon~S Du, Wei Hu, Zhiyuan Li, Ruslan Salakhutdinov, and
  Ruosong Wang.
\newblock On exact computation with an infinitely wide neural net.
\newblock In {\em NeurIPS}. \url{https://arxiv.org/pdf/1904.11955.pdf}, 2019.

\bibitem[ADH{\etalchar{+}}19b]{adhlw19}
Sanjeev Arora, Simon~S Du, Wei Hu, Zhiyuan Li, and Ruosong Wang.
\newblock Fine-grained analysis of optimization and generalization for
  overparameterized two-layer neural networks.
\newblock In {\em ICML}. \url{https://arxiv.org/pdf/1901.08584.pdf}, 2019.

\bibitem[AZLL19]{all19}
Zeyuan Allen-Zhu, Yuanzhi Li, and Yingyu Liang.
\newblock Learning and generalization in overparameterized neural networks,
  going beyond two layers.
\newblock In {\em NeurIPS}. \url{https://arxiv.org/pdf/1811.04918.pdf}, 2019.

\bibitem[AZLS19a]{als19b}
Zeyuan Allen-Zhu, Yuanzhi Li, and Zhao Song.
\newblock A convergence theory for deep learning via over-parameterization.
\newblock In {\em ICML}. \url{https://arxiv.org/pdf/1811.03962.pdf}, 2019.

\bibitem[AZLS19b]{als19a}
Zeyuan Allen-Zhu, Yuanzhi Li, and Zhao Song.
\newblock On the convergence rate of training recurrent neural networks.
\newblock In {\em NeurIPS}. \url{https://arxiv.org/pdf/1810.12065.pdf}, 2019.

\bibitem[BCMC19]{bcmc19}
Arjun~Nitin Bhagoji, Supriyo Chakraborty, Prateek Mittal, and Seraphin Calo.
\newblock Analyzing federated learning through an adversarial lens.
\newblock In {\em International Conference on Machine Learning}, pages
  634--643. PMLR, 2019.

\bibitem[Ber24]{b24}
Sergei Bernstein.
\newblock On a modification of chebyshev's inequality and of the error formula
  of laplace.
\newblock {\em Ann. Sci. Inst. Sav. Ukraine, Sect. Math}, 1(4):38--49, 1924.

\bibitem[BPSW21]{bpsw21}
Jan van~den Brand, Binghui Peng, Zhao Song, and Omri Weinstein.
\newblock Training (overparametrized) neural networks in near-linear time.
\newblock In {\em ITCS}. \url{https://arxiv.org/pdf/2006.11648.pdf}, 2021.

\bibitem[CH20]{ch20}
Francesco Croce and Matthias Hein.
\newblock Reliable evaluation of adversarial robustness with an ensemble of
  diverse parameter-free attacks.
\newblock In {\em International Conference on Machine Learning}, pages
  2206--2216. PMLR, 2020.

\bibitem[Che52]{c52}
Herman Chernoff.
\newblock A measure of asymptotic efficiency for tests of a hypothesis based on
  the sum of observations.
\newblock {\em The Annals of Mathematical Statistics}, pages 493--507, 1952.

\bibitem[CW17]{cw17c}
Nicholas Carlini and David Wagner.
\newblock Towards evaluating the robustness of neural networks.
\newblock In {\em 2017 IEEE Symposium on Security and Privacy (SP)}, pages
  39--57. IEEE, 2017.

\bibitem[DKM20]{dkm20}
Yuyang Deng, Mohammad~Mahdi Kamani, and Mehrdad Mahdavi.
\newblock Distributionally robust federated averaging.
\newblock {\em Advances in Neural Information Processing Systems},
  33:15111--15122, 2020.

\bibitem[DLL{\etalchar{+}}19]{dllwz19}
Simon~S Du, Jason~D Lee, Haochuan Li, Liwei Wang, and Xiyu Zhai.
\newblock Gradient descent finds global minima of deep neural networks.
\newblock In {\em ICML}. \url{https://arxiv.org/pdf/1811.03804}, 2019.

\bibitem[DZPS19]{dzps19}
Simon~S Du, Xiyu Zhai, Barnabas Poczos, and Aarti Singh.
\newblock Gradient descent provably optimizes over-parameterized neural
  networks.
\newblock In {\em ICLR}. \url{https://arxiv.org/pdf/1810.02054.pdf}, 2019.

\bibitem[GCL{\etalchar{+}}19]{gcl+19}
Ruiqi Gao, Tianle Cai, Haochuan Li, Cho-Jui Hsieh, Liwei Wang, and Jason~D Lee.
\newblock Convergence of adversarial training in overparametrized neural
  networks.
\newblock {\em Advances in Neural Information Processing Systems},
  32:13029--13040, 2019.

\bibitem[GSS14]{gss14}
Ian~J Goodfellow, Jonathon Shlens, and Christian Szegedy.
\newblock Explaining and harnessing adversarial examples.
\newblock {\em arXiv preprint arXiv:1412.6572}, 2014.

\bibitem[Haz16]{h16}
Elad Hazan.
\newblock Introduction to online convex optimization.
\newblock {\em Foundations and Trends in Optimization}, 2(3-4):157--325, 2016.

\bibitem[HLSY21]{hlsy21}
Baihe Huang, Xiaoxiao Li, Zhao Song, and Xin Yang.
\newblock Fl-ntk: A neural tangent kernel-based framework for federated
  learning analysis.
\newblock In {\em International Conference on Machine Learning}, pages
  4423--4434. PMLR, 2021.

\bibitem[JGH18]{jgh18}
Arthur Jacot, Franck Gabriel, and Cl{\'e}ment Hongler.
\newblock Neural tangent kernel: Convergence and generalization in neural
  networks.
\newblock In {\em Advances in neural information processing systems (NeurIPS)},
  pages 8571--8580, 2018.

\bibitem[KGB16]{kgb16b}
Alexey Kurakin, Ian Goodfellow, and Samy Bengio.
\newblock Adversarial machine learning at scale.
\newblock {\em arXiv preprint arXiv:1611.01236}, 2016.

\bibitem[LGD{\etalchar{+}}20]{lgd+20}
Xiaoxiao Li, Yufeng Gu, Nicha Dvornek, Lawrence Staib, Pamela Ventola, and
  James~S Duncan.
\newblock Multi-site fmri analysis using privacy-preserving federated learning
  and domain adaptation: Abide results.
\newblock {\em Medical Image Analysis}, 2020.

\bibitem[LHY{\etalchar{+}}19]{lhy+19}
Xiang Li, Kaixuan Huang, Wenhao Yang, Shusen Wang, and Zhihua Zhang.
\newblock On the convergence of fedavg on non-iid data.
\newblock {\em arXiv preprint arXiv:1907.02189}, 2019.

\bibitem[LJZ{\etalchar{+}}21]{ljz+21}
Xiaoxiao Li, Meirui JIANG, Xiaofei Zhang, Michael Kamp, and Qi~Dou.
\newblock Fedbn: Federated learning on non-iid features via local batch
  normalization.
\newblock In {\em International Conference on Learning Representations}, 2021.

\bibitem[LL18]{ll18}
Yuanzhi Li and Yingyu Liang.
\newblock Learning overparameterized neural networks via stochastic gradient
  descent on structured data.
\newblock In {\em NeurIPS}. \url{https://arxiv.org/pdf/1808.01204.pdf}, 2018.

\bibitem[LLC{\etalchar{+}}19]{llc+19}
Xinle Liang, Yang Liu, Tianjian Chen, Ming Liu, and Qiang Yang.
\newblock Federated transfer reinforcement learning for autonomous driving.
\newblock {\em arXiv preprint arXiv:1910.06001}, 2019.

\bibitem[LLH{\etalchar{+}}20]{llh+20}
Wei Yang~Bryan Lim, Nguyen~Cong Luong, Dinh~Thai Hoang, Yutao Jiao, Ying-Chang
  Liang, Qiang Yang, Dusit Niyato, and Chunyan Miao.
\newblock Federated learning in mobile edge networks: A comprehensive survey.
\newblock {\em IEEE Communications Surveys \& Tutorials}, 22(3):2031--2063,
  2020.

\bibitem[LSS{\etalchar{+}}20]{lsswy20}
Jason~D Lee, Ruoqi Shen, Zhao Song, Mengdi Wang, and Zheng Yu.
\newblock Generalized leverage score sampling for neural networks.
\newblock In {\em NeurIPS}, 2020.

\bibitem[LSZ{\etalchar{+}}20]{lsz+20}
Tian Li, Anit~Kumar Sahu, Manzil Zaheer, Maziar Sanjabi, Ameet Talwalkar, and
  Virginia Smith.
\newblock Federated optimization in heterogeneous networks.
\newblock In {\em Conference on Machine Learning and Systems, 2020a}, 2020.

\bibitem[MMR{\etalchar{+}}17]{mmr+17}
Brendan McMahan, Eider Moore, Daniel Ramage, Seth Hampson, and Blaise~Aguera
  y~Arcas.
\newblock Communication-efficient learning of deep networks from decentralized
  data.
\newblock In {\em Artificial Intelligence and Statistics}, pages 1273--1282.
  PMLR, 2017.

\bibitem[MMS{\etalchar{+}}18]{mmstv18}
Aleksander Madry, Aleksandar Makelov, Ludwig Schmidt, Dimitris Tsipras, and
  Adrian Vladu.
\newblock Towards deep learning models resistant to adversarial attacks.
\newblock In {\em ICLR}. \url{https://arxiv.org/pdf/1706.06083.pdf}, 2018.

\bibitem[MOSW22]{mosw22}
Alexander Munteanu, Simon Omlor, Zhao Song, and David Woodruff.
\newblock Bounding the width of neural networks via coupled initialization a
  worst case analysis.
\newblock In {\em International Conference on Machine Learning}, pages
  16083--16122. PMLR, 2022.

\bibitem[RCZ{\etalchar{+}}20]{rcz+20}
Sashank Reddi, Zachary Charles, Manzil Zaheer, Zachary Garrett, Keith Rush,
  Jakub Kone{\v{c}}n{\`y}, Sanjiv Kumar, and H~Brendan McMahan.
\newblock Adaptive federated optimization.
\newblock {\em arXiv preprint arXiv:2003.00295}, 2020.

\bibitem[RFPJ20]{rfpj20}
Amirhossein Reisizadeh, Farzan Farnia, Ramtin Pedarsani, and Ali Jadbabaie.
\newblock Robust federated learning: The case of affine distribution shifts.
\newblock {\em arXiv preprint arXiv:2006.08907}, 2020.

\bibitem[RHL{\etalchar{+}}20]{rhl+20}
Nicola Rieke, Jonny Hancox, Wenqi Li, Fausto Milletari, Holger~R Roth, Shadi
  Albarqouni, Spyridon Bakas, Mathieu~N Galtier, Bennett~A Landman, Klaus
  Maier-Hein, et~al.
\newblock The future of digital health with federated learning.
\newblock {\em NPJ digital medicine}, 3(1):1--7, 2020.

\bibitem[SKC18]{skc18}
Pouya Samangouei, Maya Kabkab, and Rama Chellappa.
\newblock Defense-gan: Protecting classifiers against adversarial attacks using
  generative models.
\newblock {\em arXiv preprint arXiv:1805.06605}, 2018.

\bibitem[SY19]{sy19}
Zhao Song and Xin Yang.
\newblock Quadratic suffices for over-parametrization via matrix chernoff
  bound.
\newblock In {\em arXiv preprint}. \url{https://arxiv.org/pdf/1906.03593.pdf},
  2019.

\bibitem[SYZ21]{syz21}
Zhao Song, Shuo Yang, and Ruizhe Zhang.
\newblock Does preprocessing help training over-parameterized neural networks?
\newblock {\em Advances in Neural Information Processing Systems}, 34, 2021.

\bibitem[SZS{\etalchar{+}}13]{szsbegf13}
Christian Szegedy, Wojciech Zaremba, Ilya Sutskever, Joan Bruna, Dumitru Erhan,
  Ian Goodfellow, and Rob Fergus.
\newblock Intriguing properties of neural networks.
\newblock In {\em arXiv preprint}. \url{https://arxiv.org/pdf/1312.6199.pdf},
  2013.

\bibitem[SZZ21]{szz21}
Zhao Song, Lichen Zhang, and Ruizhe Zhang.
\newblock Training multi-layer over-parametrized neural network in subquadratic
  time.
\newblock {\em arXiv preprint arXiv:2112.07628}, 2021.

\bibitem[TKP{\etalchar{+}}17]{tkp+17}
Florian Tram{\`e}r, Alexey Kurakin, Nicolas Papernot, Ian Goodfellow, Dan
  Boneh, and Patrick McDaniel.
\newblock Ensemble adversarial training: Attacks and defenses.
\newblock {\em arXiv preprint arXiv:1705.07204}, 2017.

\bibitem[TZT18]{tzt18}
Zhuozhuo Tu, Jingwei Zhang, and Dacheng Tao.
\newblock Theoretical analysis of adversarial learning: A minimax approach.
\newblock {\em arXiv preprint arXiv:1811.05232}, 2018.

\bibitem[WLL{\etalchar{+}}20]{wll+20}
Jianyu Wang, Qinghua Liu, Hao Liang, Gauri Joshi, and H~Vincent Poor.
\newblock Tackling the objective inconsistency problem in heterogeneous
  federated optimization.
\newblock {\em arXiv preprint arXiv:2007.07481}, 2020.

\bibitem[WTS{\etalchar{+}}19]{wts+19}
Shiqiang Wang, Tiffany Tuor, Theodoros Salonidis, Kin~K. Leung, Christian
  Makaya, Ting He, and Kevin Chan.
\newblock Adaptive federated learning in resource constrained edge computing
  systems.
\newblock {\em IEEE Journal on Selected Areas in Communications},
  37(6):1205--1221, 2019.

\bibitem[YCKB18]{yckb18}
Dong Yin, Yudong Chen, Ramchandran Kannan, and Peter Bartlett.
\newblock Byzantine-robust distributed learning: Towards optimal statistical
  rates.
\newblock In {\em International Conference on Machine Learning}, pages
  5650--5659. PMLR, 2018.

\bibitem[Zha22]{z22}
Lichen Zhang.
\newblock Speeding up optimizations via data structures: Faster search, sample
  and maintenance.
\newblock Master's thesis, Carnegie Mellon University, 2022.

\bibitem[ZHD{\etalchar{+}}20]{zhd+20}
Xinwei Zhang, Mingyi Hong, Sairaj Dhople, Wotao Yin, and Yang Liu.
\newblock Fedpd: A federated learning framework with optimal rates and
  adaptivity to non-iid data.
\newblock {\em arXiv preprint arXiv:2005.11418}, 2020.

\bibitem[ZLL{\etalchar{+}}18]{zls+18}
Yue Zhao, Meng Li, Liangzhen Lai, Naveen Suda, Damon Civin, and Vikas Chandra.
\newblock Federated learning with non-iid data.
\newblock {\em arXiv preprint arXiv:1806.00582}, 2018.

\bibitem[ZLL{\etalchar{+}}21]{zll+21}
Gaoyuan Zhang, Songtao Lu, Sijia Liu, Xiangyi Chen, Pin-Yu Chen, Lee Martie,
  and Mingyi Horesh, Lior abd~Hong.
\newblock Distributed adversarial training to robustify deep neural networks at
  scale.
\newblock 2021.

\bibitem[ZPD{\etalchar{+}}20]{zpdlsa20}
Yi~Zhang, Orestis Plevrakis, Simon~S Du, Xingguo Li, Zhao Song, and Sanjeev
  Arora.
\newblock Over-parameterized adversarial training: An analysis overcoming the
  curse of dimensionality.
\newblock In {\em NeurIPS}. arXiv preprint arXiv:2002.06668, 2020.

\bibitem[ZRSB20]{zrsb20}
Giulio Zizzo, Ambrish Rawat, Mathieu Sinn, and Beat Buesser.
\newblock Fat: Federated adversarial training.
\newblock {\em arXiv preprint arXiv:2012.01791}, 2020.

\end{thebibliography}
\section*{Checklist}

\begin{enumerate}

\item For all authors...
\begin{enumerate}
  \item Do the main claims made in the abstract and introduction accurately reflect the paper's contributions and scope?
    \answerYes{}
  \item Did you describe the limitations of your work?
    \answerYes{} We describe the limitations in conclusion in Section~\ref{sec:conclusion}.
  \item Did you discuss any potential negative societal impacts of your work?
    \answerYes{} We propose a general framework for adversarial training in federated learning setting and make theoretical analysis of it. It does not have potential negative societal impacts. See discussion in Section~\ref{sec:intro}.
  \item Have you read the ethics review guidelines and ensured that your paper conforms to them?
    \answerYes{}
\end{enumerate}

\item If you are including theoretical results...
\begin{enumerate}
  \item Did you state the full set of assumptions of all theoretical results?
    \answerYes{}
        \item Did you include complete proofs of all theoretical results?
    \answerYes{}
\end{enumerate}

\item If you ran experiments...
\begin{enumerate}
  \item Did you include the code, data, and instructions needed to reproduce the main experimental results (either in the supplemental material or as a URL)?
    \answerNA{}
  \item Did you specify all the training details (e.g., data splits, hyperparameters, how they were chosen)?
    \answerNA{}
        \item Did you report error bars (e.g., with respect to the random seed after running experiments multiple times)?
    \answerNA{}
        \item Did you include the total amount of compute and the type of resources used (e.g., type of GPUs, internal cluster, or cloud provider)?
    \answerNA{}
\end{enumerate}

\item If you are using existing assets (e.g., code, data, models) or curating/releasing new assets...
\begin{enumerate}
  \item If your work uses existing assets, did you cite the creators?
    \answerNA{}
  \item Did you mention the license of the assets?
    \answerNA{}
  \item Did you include any new assets either in the supplemental material or as a URL?
    \answerNA{}
  \item Did you discuss whether and how consent was obtained from people whose data you're using/curating?
    \answerNA{}
  \item Did you discuss whether the data you are using/curating contains personally identifiable information or offensive content?
    \answerNA{}
\end{enumerate}

\item If you used crowdsourcing or conducted research with human subjects...
\begin{enumerate}
  \item Did you include the full text of instructions given to participants and screenshots, if applicable?
    \answerNA{}
  \item Did you describe any potential participant risks, with links to Institutional Review Board (IRB) approvals, if applicable?
    \answerNA{}
  \item Did you include the estimated hourly wage paid to participants and the total amount spent on participant compensation?
    \answerNA{}
\end{enumerate}

\end{enumerate}
\fi






\end{document}